\documentclass[journal]{IEEEtran}
\usepackage{booktabs} 
\usepackage{times}
\usepackage{helvet}
\usepackage{multirow}
\usepackage{booktabs}  
\usepackage{amsmath}
\usepackage{threeparttable}  
\usepackage{makecell}
\usepackage{graphicx}
\usepackage{stfloats}
\usepackage{subfigure}
\usepackage{epstopdf}
\usepackage{algorithm}
\usepackage{algorithmic}
\usepackage{float}
\usepackage{bm}
\usepackage{fancyhdr}
\usepackage{courier}
\usepackage{subfigure} 
\usepackage{epsfig}
\usepackage{graphicx}
\usepackage{url}
\usepackage{epstopdf}
\usepackage{amssymb}
\usepackage{amsfonts}
\usepackage{amsmath}
\usepackage{ragged2e}
\usepackage{amsmath}
\usepackage{mathrsfs}
\usepackage{algorithm}
\usepackage{algorithmic}
\usepackage{multirow}
\usepackage{booktabs}
\usepackage{CJK}
\usepackage{psfrag}    
\usepackage{lineno}
\usepackage{subfig}
\usepackage[dvipsnames]{xcolor}
\usepackage{thmtools}
\usepackage{enumitem}
\usepackage{bm}
\usepackage{pgfplots}
\usepackage{verbatim}
\usepackage{float}
\usepackage{mathtools}
\usepackage{cuted}
\usepackage{capt-of}
\usepackage{color}
\usepackage{cite} 

\begin{document}

%
\title{Semantic-aware Modular Capsule Routing for Visual Question Answering}
%
%
%
\author{
	Yudong~Han,~Jianhua~Yin,~\IEEEmembership{Member,~IEEE,}~Jianlong~Wu,~\IEEEmembership{Member,~IEEE,}\\~Yinwei~Wei,~Liqiang~Nie,~\IEEEmembership{Senior Member,~IEEE,}\\
\thanks{This work is supported by the National Natural Science Foundation of China, No.:U1936203, No.:62006142; the Shandong Provincial Natural Science Foundation, No.:ZR2019JQ23; }

\IEEEcompsocitemizethanks{\IEEEcompsocthanksitem Yudong Han, Jianhua Yin, and Jianlong Wu are with the School of Computer Science and Technology, Shandong University, Qingdao, 266237, China. (e-mail: hanyudong.sdu@gmail.com; jhyin@sdu.edu.cn; jlwu1992@sdu.edu.cn)
\IEEEcompsocthanksitem Yinwei Wei is currently a research fellow with
NExT++, National University of Singapore. (e-mail: weiyinwei@hotmail.com)
\IEEEcompsocthanksitem Liqiang Nie is a professor with Harbin Institute of Technology (Shenzhen). (e-mail: nieliqiang@gmail.com)
}}

\markboth{IEEE TRANSACTIONS ON IMAGE PROCESSING, 2022}%
{Shell \MakeLowercase{\textit{et al.}}: Bare Demo of IEEEtran.cls for IEEE Journals}

\maketitle


\begin{abstract}
Visual Question Answering (VQA) is fundamentally compositional in nature, and many questions are simply answered by decomposing them into modular sub-problems. The recent proposed Neural Module Network (NMN) employ this strategy to question answering, whereas heavily rest with off-the-shelf layout parser or additional expert policy regarding the network architecture design instead of learning from the data. These strategies result in the unsatisfactory adaptability to the semantically-complicated variance of the inputs, thereby hindering the representational capacity and generalizability of the model. To tackle this problem, we propose a \textbf{S}emantic-aware mod\textbf{U}lar ca\textbf{P}sul\textbf{E} \textbf{R}outing framework, termed as SUPER, to better capture the instance-specific vision-semantic characteristics and refine the discriminative representations for prediction. Particularly, five powerful specialized modules as well as dynamic routers are tailored in each layer of the SUPER network, and the compact routing spaces are constructed such that a variety of customizable routes can be sufficiently exploited and the vision-semantic representations can be explicitly calibrated. We comparatively justify the effectiveness and generalization ability of our proposed SUPER scheme over five benchmark datasets, as well as the parametric-efficient advantage. It is worth emphasizing that this work is not to pursue the state-of-the-art results in VQA. Instead, we expect that our model is responsible to provide a novel perspective towards architecture learning and representation calibration for VQA. 
\end{abstract}


\section{Introduction}
\IEEEPARstart{V}{isual} Question Answering (VQA) has emerged as a challenging interdisciplinary research problem over the past few years. It targets at correctly answering a natural language question about a given image. With the increasing prosperity of both computer vision and natural language processing communities, numerous well-designed approaches~\cite{DBLP:conf/cvpr/AndreasRDK16, DBLP:conf/cvpr/CadeneBCT19} have been devised towards this task. Thereinto, the mainstreaming methods~\cite{DBLP:conf/iccv/HuARDS17, DBLP:conf/eccv/HuADS18, DBLP:conf/cvpr/ShiZL19} design an adaptive network to improve the representational capacity and generalizability of the model. In these methods, they reason the answer by parsing questions into linguistic substructures and assembling question-specific modular networks from collections of function-specific modules. 

\begin{figure}
    \centering
    \includegraphics[width=0.48\textwidth]{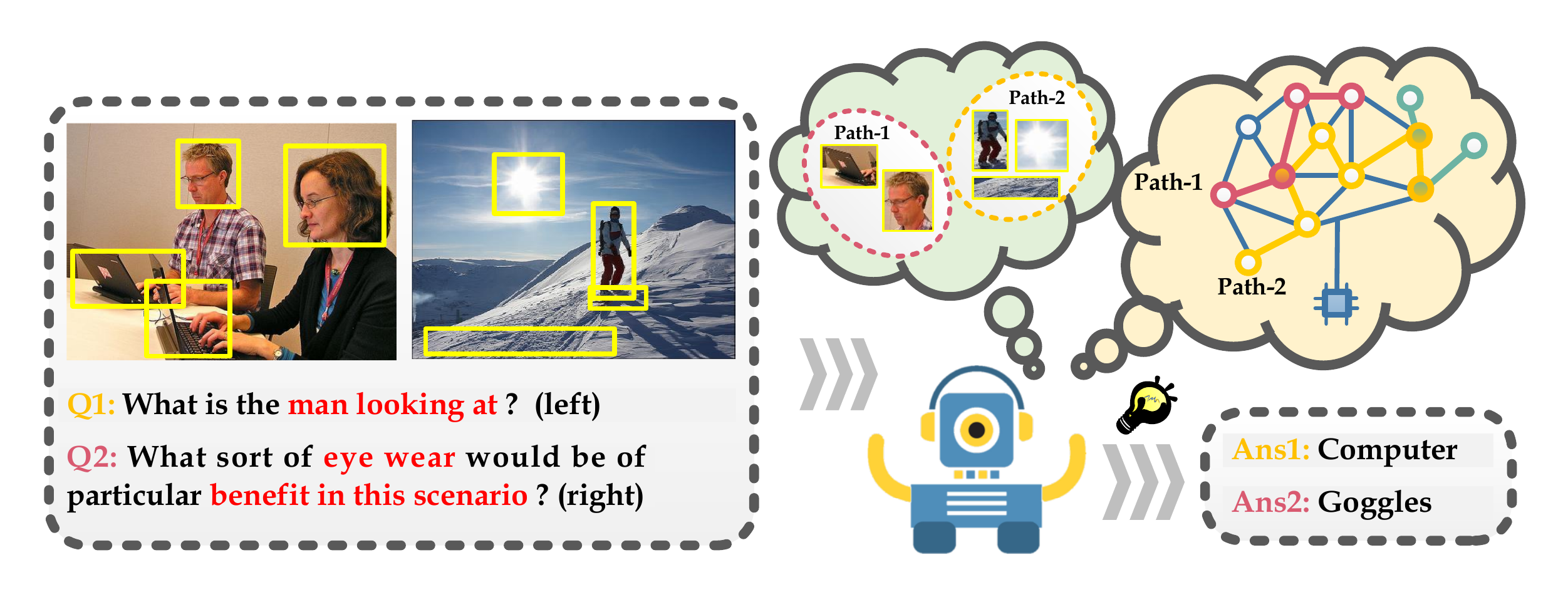}
    \vspace{-0.5em}
    \caption{Given the image-question pairs embodying different visual scenes, our proposed scheme is inclined to activate different routes. Moreover, the semantically-complicated image-question pairs (Q2) are generally handled with more sophisticated reasoning routes, which simulates the decision-making process of our human brain.}
    \label{fig:introduction}
    \vspace{-2em}
\end{figure}

Despite their promising performance, these adaptive-designed approaches still exhibit the following fundamental limitations: 1) they rely on vulnerable off-the-shelf language parsers or expert policies which are not specially designed for the language and vision task, and the network layout is guided by these additional supervision rather than learning from the input data. Therefore, they lack the adaptability to diverse vision-semantic distributions in the real-world settings and hence may be invalid. 2) Beyond that, the devised composable function-specific modules in these methods are restricted to reasoning the simple scene from synthesized datasets with low variability, such as CLEVR~\cite{DBLP:conf/cvpr/JohnsonHMFZG17}, which might not be well adapted to more complicated visual scene. 

We argue that, the key to tackling aforementioned issues is to unify the compact network structure and powerful modules into a dynamic data-driven setting. As illustrated in Figure~\ref{fig:introduction}, given a question that can be answered simply by visual recognition, the model should accurately give the answer ``computer'' without complex reasoning routes. As to the second question which needs more delicate reasoning, it requires the model to first infer the visual scene of skiing according to the contextual information, such as ``sun'', ``people'', and ``snow'', and then leverage the related knowledge (e.g., goggles, used for, skiing) as an auxiliary information to reply. In the ideal scenario, the model should be capable of choosing specific processing routes when handling diverse inputs for the needs of flexible and explicit reasoning.

In light of this, inspired by the $\textit{dynamic routing}$ in \textit{CapsNet}~\cite{DBLP:conf/nips/SabourFH17}, we devise a \textbf{S}emantic-aware mod\textbf{U}lar ca\textbf{P}sul\textbf{E} \textbf{R}outing network, abbreviated as SUPER, to better accommodate the semantically-complicated variance of different image-question pairs and discriminate the instance-specific vision-semantic representation. As shown in Figure~\ref{fig:framework} (a), the SUPER network treats image region features as underlying capsules and reasons the visual scene with respect to a question via iterating through a shared SUPER layer. In terms of the SUPER layer, three reciprocal components are sequentially deployed for delicate capsule modulation, thereby obtaining the discriminative representations for answering. To be specific, the capsules first undergo the semantic-aware modular routing, wherein five specialized modules are configured in parallel for powerful visual modulation in VQA, as shown in Figure~\ref{fig:framework} (c). Meanwhile, a dynamic router is integrated into each module to derive the data-dependent paths for modulation. Beyond that, we construct the dense-connected paths among these modules in two adjacent time iterations such that a variety of customizable routes can be sufficiently exploited. After processing by the modular routing, these modulated capsules are aggregated by the gating agreements, which further determines the importance of these capsules to the answer. Beyond that, to further preserve and rectify the discriminative clues favorable for answering, a memory-augmented component is intentionally employed.


To evaluate our proposed model, we conduct extensive experiments on five publicly accessible benchmark datasets, namely, VQA-v2, VQA-CP v2, OK-VQA, FVQA, and Visual7W. By comparing with numerous state-of-the-art baselines, we demonstrate the effectiveness and generalization capability of our proposed model. 

In summary, the contributions of this work are three-fold: 
\begin{itemize}
	\item We present a semantic-aware modular capsule routing scheme towards VQA, which is the first attempt to integrate the modular-based dynamic routing into the capsule network.
	\item We elaborately devise three reciprocal components, namely, including semantic-aware modular routing, gating agreements, and memory reactivation, which guide the model to dynamically select the specific paths according to the inputs and perform delicate capsule modulation via iterations.
	\item Regarding the modular architecture, we tailor five specialized modules and integrate a router in each module for flexible exploration of the modulation pattern.
\end{itemize}

\section{Related Work}
\subsection{Visual Question Answering} The VQA task requires co-reasoning over both the visual content and question to infer the correct answer, which can be categorized into two groups: conventional VQA and knowledge-required VQA.

For conventional visual questions, the pioneer studies embed the visual and question features into a common space by various multimodal embedding techniques~\cite{DBLP:conf/emnlp/FukuiPYRDR16, DBLP:conf/aaai/Ben-younesCTC19}. For instance, the authors in~\cite{DBLP:conf/emnlp/FukuiPYRDR16} first proposed Multimodal Compact Bilinear pooling (MCB), which utilizes bilinear pooling to enhance the multi-modal representation, yet it brings high computational cost. Compared to MCB, Multi-modal Factorized Bilinear pooling (MFB)~\cite{DBLP:conf/iccv/YuY0T17} obtains more compact features using the Hardamard product operation and achieves strong performance with lower computational cost. Different from above embedding-based models, Anderson \textit{et al.}~\cite{DBLP:conf/cvpr/00010BT0GZ18} first introduced a Bottom-Up and Top-Down attention mechanism, which assigns a distinctive attention distribution on detected image objects for efficiently answering. This technique has been widely applied in recent studies ~\cite{DBLP:conf/nips/KimJZ18, DBLP:conf/cvpr/Yu0CT019}, boosting the performance of a series of VQA models. For instance, BAN~\cite{DBLP:conf/nips/KimJZ18} adopts the bilinear attention between the image objects and question words, which could fuse two modalities seamlessly. Moreover, DFAF~\cite{DBLP:conf/cvpr/GaoJYLHWL19} introduces the self-attention and cross-attention layer into VQA, where the intra- and inter-modality interactions are both effectively modeled.
Orthogonal to these attention-based studies, a series of delicate relation-based models~\cite{DBLP:conf/nips/Norcliffe-Brown18, DBLP:conf/iccv/LiGCL19} are proposed to enhance the contextual visual representation by exploring the visual relations between image objects. For example, in~\cite{DBLP:conf/nips/Norcliffe-Brown18}, a graph learner is introduced to reason the relation representation through the pairwise attention modeling and spatial graph convolution. To further model multi-type inter-object interactions, ReGAT~\cite{DBLP:conf/iccv/LiGCL19} considers both explicit and implicit relations to enrich the visual representation.  

Other strands of research~\cite{DBLP:journals/pami/WangWSDH18, DBLP:conf/cvpr/MarinoRFM19} have attempted to leverage information beyond the image-question pair, such as retrieving basic factual knowledge or external commonsense knowledge to answer the questions. Initially, Wang \textit{et al.} ~\cite{DBLP:journals/pami/WangWSDH18} introduced a fact-based VQA task, which provides a close-domain knowledge of facts and associates each question with a supporting-fact. Recently, Marino \textit{et al.}~\cite{DBLP:conf/cvpr/MarinoRFM19} provided an open-domain benchmark, called OK-VQA, which requires the model to introduce knowledge from the open knowledge sources~\cite{DBLP:conf/aaai/SpeerCH17} to assist the answering. With regard to these two datasets, numerous models ~\cite{DBLP:conf/nips/NarasimhanLS18, DBLP:conf/mm/Li0020} thrive around and have achieved considerable performance improvement.
 \vspace{-2ex}
\subsection{Neural Module Networks} 
There are a series of studies~\cite{DBLP:conf/nips/SongLJL00020, DBLP:conf/iccv/HuARDS17, DBLP:conf/cvpr/WuNKRDGF18, DBLP:conf/mm/YuCYWTT20} delving in exploiting the network architectures for better generalized capability towards VQA. Different from Neural Architecture Search (NAS)~\cite{DBLP:conf/mm/YuCYWTT20} with static inference, the recent proposed Neural Module Network (NMN) ~\cite{DBLP:conf/cvpr/AndreasRDK16} constructs the instance-specific network conditioned on the syntactic structure of the question. In a NMN model, every input is associated with a layout that provides a template for assembling the network structure from a collection of shallow function-specific modules. These compositional modules can be jointly trained across multiple structures to implement task-specific reasoning and answering. The prior work on NMN~\cite{DBLP:conf/cvpr/AndreasRDK16} uses fixed rule-based layout generated from the off-the-shelf language parser. Later work~\cite{DBLP:conf/iccv/HuARDS17} attempts to specify an optimal layout predictor for a modular neural network using the reverse Polish notation. 

Different from current related studies, our model adopts the modular routing strategy to compose the powerful specialized modules into a dynamic network for VQA, which could derive the data-dependent forward path according
to the distribution of each image-question input.
\subsection{Capsule Networks} Recently, capsule networks have been actively applied to many fields, such as graph network, object detection, and recommendation. Sabour \textit{et al.}~\cite{DBLP:conf/nips/SabourFH17} first introduced the idea of capsules in neural network towards the classification task, where the capsules can represent various properties of certain object. Specifically, an underlying capsule can be activated if its represented property are pivotal to the prediction, and it can be further routed to the overlying capsules iteratively to solve the problem. Due to the advantage of local pattern activation and the combination of capsule networks, some variants and applications of it have been explored recently. For instance, Gu \textit{et al.}~\cite{DBLP:conf/aaai/Gu21} introduced a graph capsule network for the object recognition task, where the routing part is replaced with a multi-head attention-based graph pooling method for better local relational modeling. In the field of VQA, Zhou \textit{et al.}~\cite{DBLP:conf/aaai/ZhouJSSC19} revealed the possibility of performing multi-step attentions in a dynamic capsule network towards robust and efficient attention modeling. 
However, how to better adapt the capsule network to the VQA task still remains a critical challenge.
\vspace{-1em}

\section{Our Proposed Method}

As shown in Figure 2 (a), our proposed SUPER model is comprised of
three main components: Feature Representation (Section 3.1), SUPER Network (Section 3.2), and Answer Reasoning (Section 3.4). In what follows, we elaborate these three components sequentially.
\subsection{Feature Representation}
\subsubsection{Image Embedding}
Following~\cite{DBLP:conf/cvpr/00010BT0GZ18}, we extract image features via a pre-trained Faster-RCNN model~\cite{DBLP:conf/nips/RenHGS15}. Concretely, we first select $N$ bounding boxes with the highest class confidence scores for each image. The obtained visual feature matrix is denoted as $\mathbf{F} \in \mathbb{R}^{d_{v} \times N}$, where each column represents one region feature. Thereafter, we feed $\mathbf{F}$ into a fully-connected layer, obtaining the final visual representation $\mathbf{V} \in \mathbb{R}^{d \times N}$. 
\subsubsection{Question Embedding}
For the question embedding, we adopt GRU~\cite{DBLP:conf/emnlp/ChoMGBBSB14} followed by a self-attention layer~\cite{DBLP:conf/nips/VaswaniSPUJGKP17} to generate the context-aware words embedding matrix $\mathbf{\hat{Q}} \in \mathbb{R}^{d \times L}$, where $L$ represents the length of the question, and each column of $\mathbf{\hat{Q}}$ is the representation of each word. Considering that each word conveys distinctive contextual semantic information and weighs differently, we hereby use the local attention mechanism to regulate each word feature as follows, 
\begin{equation}
    \label{eq:0}
    \mathbf{Q} = \mathbf{\hat{Q}} \ \mathbf{diag}(\delta_{a}(\mathbf{W}_{\hat{Q}} \mathbf{\hat{Q}})),
\end{equation}
where $\mathbf{diag}(\cdot)$ denotes diagonalization operation on a vector, $\delta_{a}(\cdot)$ is the softmax activation, and $\mathbf{W}_{\hat{Q}} \in \mathbb{R}^{1 \times d}$ represents the attention mapping function.
\subsubsection{Knowledge Embedding}
For the knowledge-required image-question pairs, we associate the key entities appearing in the image and question to large-scale knowledge graph, and obtain the related knowledge embedding $\textbf{K} \in \mathbb{R}^{d \times K}$ to adapt the knowledge-required VQA, where $K$ represents the number of extracted knowledge facts for each image-question pair.
Our knowledge embedding stage is three-fold: \textbf{1) Entity Recognization.} We recognize the tokens $\mathcal{C}_{q}$ in the given question using the spacy library\footnote{https://github.com/explosion/spaCy}, and obtain the semantic labels $\mathcal{C}_{o}$ of the detected image objects via a pretrained Faster-RCNN~\cite{DBLP:conf/nips/RenHGS15}. Take the right image in Figure 1 of the original manuscript as an example, in the given question ``What sort of eye wear would be of particular benefit in this scenario?'', the extracted $\mathcal{C}_{q}$ would be $\left \{ sort, eye, wear, eyewear, benefit, scenario \right \}$, and $\mathcal{C}_{o}$ can be represented as $\left \{ sun, people, snow, skiing park, etc. \right \}$. \textbf{2) Subgraph Construction via Path Retrieving.} Before constructing the subgraph, we first briefly introduce our target knowledge graph $\textit{ConceptNet}$~\cite{DBLP:conf/aaai/SpeerCH17}. $\textit{ConceptNet}$ can be regarded as a large set of triplets $(h, r, t)$, like (cake, usedFor, wedding), where $h$ and $t$ denote head and tail entities, respectively, and $r$ is a certain relation type. Following~\cite{DBLP:conf/emnlp/LinCCR19}, we roughly divide the relations into 17 types to increase the density of the knowledge graph. Based on the $\textit{ConceptNet}$, we construct the subgraph by finding the knowledge paths among the mentioned entities $\mathcal{C}_{q} \cup \mathcal{C}_{o}$. To be specific, for each question entity $e_{i} \in \mathcal{C}_{q}$ and image entity $e_{j} \in \mathcal{C}_{o}$, we find the paths between them that are shorter than $R$ relations\footnote{To improve the efficiency of retrieval, we set $R=2$.}. \textbf{3) Knowledge Path Encoding.} After obtaining the paths, we generate the corresponding knowledge representations $\mathbf{K} \in \mathbb{R}^{d_{k} \times K}$ for each image-question pair via a pretrained model used in ~\cite{DBLP:conf/aaai/MalaviyaBBC20}, where $K$ is the number of the paths (i.e., the number of knowledge representations). 
\subsection{SUPER Network} 
We propose the SUPER network, a modular routing architecture that iterates through a shared SUPER layer. For conventional VQA, the SUPER layer takes as input the visual features $\{\textbf{O}_{\rightarrow m}^{(t-1)}\}_{m=1}^{M}$, question words embedding $\mathbf{Q}$, and memory $\mathbf{U}^{(t-1)}  \in \mathbb{R}^{d \times N}$ in each iteration, where $\textbf{O}_{\rightarrow m}^{(t-1)} = \left[ \textbf{O}_{1 \rightarrow m}^{(t-1)}, \cdots, \textbf{O}_{M \rightarrow m}^{(t-1)} \right ] \in \mathbb{R}^{M \times d \times N}$ denotes the output of other modules to the $m$-th module in the ($t$-1)-th iteration, and each element is initialized as $\mathbf{V}$, i.e., $\textbf{O}_{m^{*} \rightarrow m}^{(0)} = \mathbf{V}$. For knowledge-required VQA, the knowledge embedding $\mathbf{K}$ is additionally required. The outputs of the $t$-th layer are represented as $\{\textbf{O}_{\rightarrow m}^{(t)}\}_{m=1}^{M}$ and $\mathbf{U}^{(t) } \in \mathbb{R}^{d \times N}$.
\subsubsection{SUPER Layer} 
\begin{figure*}
 \hspace*{-1.5em}
 \includegraphics[width=190mm]{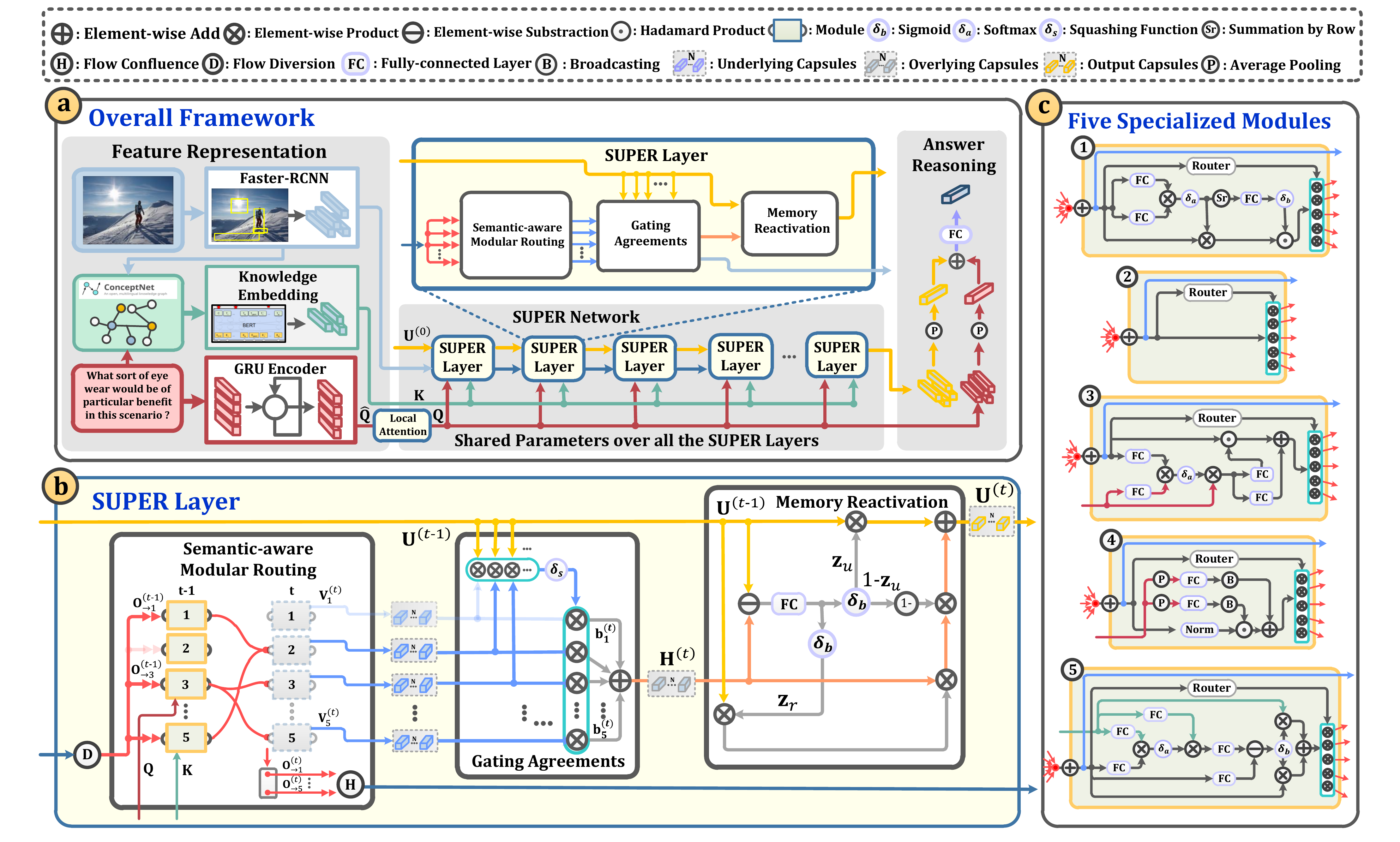}
 \vspace{-2ex}
 \caption{Schematic illustration of our proposed SUPER model. Part (a) curates the overall framework of our SUPER model. This SUPER model leverages the power of the SUPER network, a VQA architecture that iterates through a SUPER layer to reason the scene with respect to a question. Five specialized modules are devised in the semantic-aware modular routing, and the dense-connected paths are constructed among these modules in two adjacent time iterations to form a routing space, and we only mark the activated paths in red. Part (b) illustrates the implementation details of the SUPER Layer. Part (c) denotes the design details of five function-specific modules.}
 \vspace{-2ex}
 \label{fig:framework}
\end{figure*}

As shown in Figure~\ref{fig:framework} (b), there are three key parts in the SUPER layer: Semantic-aware Modular Routing, Gating Agreements, and Memory Reactivation.

\textbf{Semantic-aware Modular Routing}. In terms of modular routing, a series of specialized modules are tailored to implement delicate visual modulation, which will be elaborated in subsection 3.3. To fully explore the respective advantages of each module, they are intentionally configured in a parallel manner. Besides, to derive the instance-specific modulation routes, we construct the dense-connected paths between these modules in two adjacent time iterations while provide a router for each module. Owing to this, each module has the chance to receive the output of all the modules in the last iteration, improving the flexibility and variety of the routing process. Moreover, considering that capsules embodying different semantic information tend to activate different routes for modulation, we adopt a $\textit{data-dependent}$ vector as the router,
\begin{equation}
   \label{eq:1}
   \begin{aligned}
   \mathbf{G}_{m}^{(t-1)} &= Router(\mathbf{V}_{m}^{(t-1)}) \\
   & =\delta_{b}(\mathbf{W}_{g}(\mathcal{P}(\mathbf{V}_{m}^{(t-1)}))),
   \end{aligned}
\end{equation}  
where $\mathbf{V}_{m}^{(t-1)} \in \mathbb{R}^{d \times N}$ denotes the inputs of the $m$-th module in the $t$-th iteration, $\mathcal{P}(\cdot)$ represents the operation of average pooling, $\mathbf{W}_{g} \in \mathbb{R}^{M \times d}$ is the learnable weight, $\delta_{b}(\cdot)$ is the sigmoid activation, and $M=5$ is the number of specialized modules. In $\mathbf{G}_{m}^{(t-1)} \in \mathbb{R}^{M}$, each element $g_{m \rightarrow m^{*}}^{(t-1)} \in (0, 1)$ represents the output control gate of the $m$-th module in the ($t$-1)-th iteration, and the output from the $m$-th module to other modules can be calculated accordingly, 
\begin{equation}
   \label{eq:2}
   \mathbf{O}_{m \rightarrow m^{*}}^{(t)} = g_{m \rightarrow m^{*}}^{(t-1)} \mathcal{R}_{m}(\mathbf{V}_{m}^{(t-1)}),
\end{equation}
where $\mathcal{R}_{m}(\cdot)$ represents the visual modulation function in the $m$-th module, and $m^{*}$ denotes the index of other module. Through this data-driven routing mechanism, the modulated capsules which are aggregated to the $m$-th module in the next iteration can be summed by the following equation,
\begin{equation}
   \label{eq:3}
    \mathbf{V}_{m}^{(t)} = \sum_{m^{*}=1}^{M} \mathbf{O}_{m^{*} \rightarrow m}^{(t)}.
\end{equation}

\textbf{Gating Agreements.} Intuitively, the capsules modulated by different modules should be assigned with distinctive weights to the predicted answer. This mechanism could enable our model to follow the considerable capsules with interest, thereby improving the discriminability of representations. To achieve this, we deploy the gating agreements followed by the modular routing process, which shares similar philosophy with the \textit{routing-by-agreement} mechanism in $\textit{CapsNet}$ ~\cite{DBLP:conf/nips/SabourFH17},
\begin{equation}
    \label{eq:4}
    \mathbf{H}^{(t)} = \sum_{m=1}^{M}   \mathbf{V}_{m}^{(t)} \ \mathbf{diag}(\mathbf{c}_{m}^{(t)}),
\end{equation}
where $\mathbf{diag}(\cdot)$ returns a square diagonal matrix with the elements of a vector, and $\mathbf{c}^{(t)}_{m} \in \mathbb{R}^{N}$ is the coupling vector, which is calculated by $\mathbf{c}^{(t)}_{m}=\delta_{a}(\mathbf{b}_{m}^{(t)})$, where $\delta_{a}(\cdot)$ is the softmax activation. $\mathbf{b}_{m}^{(t)} \in \mathbb{R}^{N}$ is the log prior probability that the underlying capsules should be coupled to the output capsules, which is initialized with a zero vector and then updated through dynamic interactions between these underlying capsules $\mathbf{V}_{m}^{(t)}$ and preceding output capsules $\mathbf{U}^{(t-1)}$,
\begin{equation}
    \label{eq:5}
    \mathbf{b}_{m}^{(t)} = \mathbf{b}^{(t-1)}_{m} + \delta_{s}(\mathbf{V}_{m}^{(t)} \odot \mathbf{U}^{(t-1)}),
\end{equation}
where $\delta_{s}(\cdot) = \frac{\Vert \cdot \Vert_{2}^{2}}{1 + \Vert \cdot \Vert_{2}^{2}} \in (0, 1)$ is used to calculate the activation rate of each capsule in the matrix\footnote{When the capsule vector $\mathbf{x}$ approaches $\mathbf{0}$, its activation rate $\delta{(\mathbf{x})}$ also approaches 0. Otherwise, $\delta{(\mathbf{x})}$ approaches 1.}, and $\odot$ denotes the element-wise multiplication operation.

\textbf{Memory Reactivation.}
To select the discriminative clues and filter out the trivial ones, we further employ two gates to regulate the update of capsules and gradually generate the representation for the whole visual scene,
\begin{equation}
    \label{eq:6}
    \left\{ \begin{array}{ll} 
     \mathbf{U}^{(t)} = (\mathbf{1} - \mathbf{z}_{u}) \odot \mathbf{U}^{(t-1)} +  \mathbf{z}_{u} \odot \mathbf{U}^{*}, \\ 
     \mathbf{U}^{*} = \mathbf{W}_{z} \ (\mathbf{z}_{r} \odot \mathbf{U}^{(t-1)}) + \mathbf{W}_{h} \mathbf{H}^{(t)},
\end{array} \right.
\end{equation}
where $\mathbf{W}_{r} \in \mathbb{R}^{d \times d}$ and $\mathbf{W}_{h} \in \mathbb{R}^{d \times d}$ are learnable weights, $\mathbf{z}_{r} \in \mathbb{R}^{d \times N}$ and $\mathbf{z}_{u} \in \mathbb{R}^{d \times N} $ respectively denote the reset gate and update gate, which decide what content to be discarded and how much the output capsules update their memory based on the difference between the overlying capsules $\mathbf{H}^{(t)}  \in \mathbb{R}^{d \times N}$ and $\mathbf{U}^{(t-1)} \in \mathbb{R}^{d \times N}$. In light of this, we formulate these two gates as the following forms,
\begin{equation}
    \label{eq:7}
    \left\{ \begin{array}{ll} 
    \mathbf{z}_{u} = \delta_{b}(\mathbf{W}_{u}(\mathbf{H}^{(t)} - \mathbf{U}^{(t-1)})), \\ 
     \mathbf{z}_{r} = \delta_{b}(\mathbf{W}_{r}(\mathbf{H}^{(t)} - \mathbf{U}^{(t-1)})),
\end{array} \right.
\end{equation}
where $\mathbf{W}_{u} \in \mathbb{R}^{d \times d}$ and $\mathbf{W}_{r} \in \mathbb{R}^{d \times d}$ are learnable weights.


For ease of exposition, we summarize the inference procedure of the SUPER network in Algorithm 1.
\begin{algorithm}[!htb]
 \caption{The Inference Procedure of the SUPER Network.}
 \label{algorithm}
 \begin{algorithmic}[1]
  \REQUIRE visual feature $\mathbf{V}$; question words feature $\mathbf{Q}$; knowledge feature $\mathbf{K}$; the number of modules $M$; the number of iterations $T$ \\
  \ENSURE the output capsules $\mathbf{U}^{(T)}$
  \STATE Initialize $\mathbf{U}^{(0)}$ as $\mathcal{B}(\mathcal{P}(\mathbf{Q}))$, where $\mathcal{B}(\cdot)$ and $\mathcal{P}(\cdot)$ denote the operation of broadcasting and average pooling, respectively.
  \STATE Initialize $\mathbf{H}^{(0)}$ as $\mathbf{0}$.
  \FOR{$m$ in $M$ interations}
  \STATE Initialize $\mathbf{V}_{m}^{(0)}$ as $\mathbf{V}$.
  \ENDFOR
  \FOR{$t$ in $T$ interations}
  \FOR{$m$ in $M$ interations}
  \STATE Obtain $\mathbf{G}_{m}^{(t-1)}$ by Eq. (2).
  \STATE Obtain the underlying capsules $\mathbf{V}_{m}^{(t)}$ by Eq. (4).
  \STATE Update the coupling coefficients $\mathbf{b}_{m}^{(t)}$ by Eq. (6).
  \ENDFOR
  \STATE Obtain the overlying capsules $\mathbf{H}^{(t)}$ by Eq. (5).
  \STATE Update the output capsules $\mathbf{U}^{(t)}$ by Eq. (7).
  \ENDFOR
  \STATE \textbf{return} $\mathbf{U}^{(T)}$
 \end{algorithmic}
\end{algorithm}

\subsubsection{SUPER Network} 
The SUPER network mimics a simple form of iterative reasoning by leveraging the power of the SUPER layer. As we can see from Figure~\ref{fig:framework} (a), the multimodal representation is progressively refined through multiple steps. More specifically, for each step $t\in [1,T]$, where $T$ is the total number of iterations, a SUPER layer processes and updates the representation for the whole visual scene as follows,
\begin{equation}
    \label{eq:8}
    \textbf{U}^{(t)},  \{\mathbf{O}_{\rightarrow m}^{(t)}\}_{m=1}^{M} = \mathrm{SUPER}_{Layer}(\mathbf{U}^{(t-1)}, \mathbf{Q}, \mathbf{K},  \{\mathbf{O}_{\rightarrow m}^{(t-1)}\}_{m=1}^{M}).
\end{equation}

At the ($t$-1)-th step $(t>1)$, the SUPER layer first curates five specialized modules to receive different semantic signals (i.e., $\mathbf{Q}$ and $\mathbf{K}$) as inputs to modulate the capsules. After agreement execution and memory reactivation, the output capsules $\textbf{U}^{(t)}$ for answering are explicitly calibrated. More crucially, the parameters are shared over the SUPER layer, which enables compact parametrization and satisfactory generalization.

\subsection{Specialized Modules for VQA} 
To strengthen the representational capacity of our SUPER network in VQA, we intentionally devise five specialized modules where each can solve one task or implement one modulation strategy. These modules can be dynamically assembled into a powerful deep network at the mercy of the routing policy, whereby flexibly reason the semantically-complicated visual scene. Formally, they can be briefly formulated as follows,
\begin{equation}
    \label{eq:9}
     \mathcal{R}_{m}(\mathbf{V}_{m}^{(t)}) = 
    \left\{ \begin{array}{ll} 
    \mathcal{F}_{m}(\mathbf{V}_{m}^{(t)}), \ m=1 \ \mathrm{or} \ 2 \\
    \mathcal{F}_{m}(\mathbf{V}_{m}^{(t)}, \mathbf{Q}), \ m=3 \ \mathrm{or} \ 4 \\
    \mathcal{F}_{m}(\mathbf{V}_{m}^{(t)}, \mathbf{K}), \ m=5.
\end{array} \right.
\end{equation}

According to the task adaptability, these modules can be further categorized into two groups: modules for conventional VQA and modules for knowledge-required VQA. In what follows, we elaborate the implement details of these two groups of modules, respectively.
\subsubsection{Specialized Module for Conventional VQA} 
\ 

\textbf{Focal Context-aware Enhanced Module.} To enhance the contextual representation of these capsules and reason the whole visual scene, we devise a focal context-aware module. Specifically, we first calculate the relative comparability between capsules,
\begin{equation}
    \label{eq:10}
    \mathbf{D}_{v, v}^{(t)} = (\mathbf{W}_{1}\mathbf{V}_{m}^{(t)})^\top (\mathbf{W}_{2}\mathbf{V}_{m}^{(t)}),
\end{equation}
where the $i$-th vector in $\mathbf{D}_{v, v}^{(t)} \in \mathbb{R}^{N \times N}$  measures the relevance of other capsules to the $i$-th capsule. Intuitively, the more relevant one capsule is to others, the more important it is. Based on this, we attempt to leverage the correlation information embodied in $\mathbf{D}_{v, v}^{(t)}$ to derive the soft importance mask for each capsule and further regulate the capsules, 
\begin{equation}
    \label{eq:11}
    \left\{ \begin{array}{ll} 
    \mathcal{R}_{1}(\mathbf{V}_{m}^{(t)}) = \delta_{a}(\mathbf{D}_{v, v}^{(t)}) ( \mathbf{V}_{m}^{(t)} \ \mathbf{diag}(\mathbf{g}_{v})), \\
     \mathbf{g}_{v} = \delta_{b}(\mathbf{W}_{3} \mathcal{S}( \mathbf{D}_{v, v}^{(t)})),
\end{array} \right.
\end{equation}
where $\mathcal{S}(\mathbf{D}_{v, v}^{(t)})  \in \mathbb{R}^{N}$, $\mathbf{g}_{v} \in \mathbb{R}^{N}$, $\mathcal{S}(\cdot)$ denotes the operation of summing by column, $\mathbf{W}_{x} \in \mathbb{R}^{d \times d} \ (x=1,2)$ and $\mathbf{W}_{3} \in \mathbb{R}^{N \times N}$ are learnable weights. 

\textbf{Identity Retainer Module.}
In the process of semantic-aware modular routing, complicated modulation operations may not be necessary, especially for the simpler image-question pairs. Inspired by this, we adopt a retainer module to simplify the information flow and alleviate the gradient vanishing problem. It can be formally formulated as $ \mathcal{R}_{2}(\mathbf{V}_{m}^{(t)})=\mathbf{V}^{(t)}_{m}$.

\textbf{Global Reduction Module.}
Global question feature conveys high-level semantic indications for better visual scene understanding. Inspired by LayerNorm~\cite{DBLP:journals/corr/BaKH16}, we introduce a novel semantic modulation method, wherein two modulation vectors $\mathbf{a}$ and $\bm{\eta}$ are derived by the global question features. Based on these two semantic vectors, the capsules are modulated as follows,
\begin{equation}
    \label{eq:12}
    \left\{ \begin{array}{ll} 
     \mathcal{R}_{3}(\mathbf{V}_{m}^{(t)}) = \mathbf{a} \odot \frac{\mathbf{V}^{(t)}_{m}-\mu(\mathbf{V}^{(t)}_{m})}{\sigma(\mathbf{V}^{(t)}_{m})} + \bm{\eta}, \\
     \mathbf{a} = \delta_{b}(\mathbf{W}_{a}\mathbf{\mathcal{B}(\mathcal{P}(\mathbf{Q}))}+\mathbf{B}_{a}), \\
     \bm{\eta} = \delta_{b}(\mathbf{W}_{\eta}\mathbf{\mathcal{B}(\mathcal{P}(\mathbf{Q}))}+\mathbf{B}_{\eta}),
\end{array} \right.
\end{equation}
where $\mathbf{a} \in \mathbb{R}^{d \times N}$, $ \bm{\eta} \in \mathbb{R}^{d \times N}$, $\mu(\cdot)$ and $\sigma(\cdot)$ respectively calculate the mean and variance of each capsule in $\mathbf{V}_{m}^{(t)}$, $\mathcal{B}(\cdot)$ and $\mathcal{P}(\cdot)$ respectively denote the operation of broadcasting and average pooling, $\mathbf{W}_{a} \in \mathbb{R}^{d \times d}$, $\mathbf{W}_{\eta} \in \mathbb{R}^{d \times d}$, $\mathbf{B}_{a} \in \mathbb{R}^{d \times d}$, and $\mathbf{B}_{\eta} \in \mathbb{R}^{d \times d}$ are learnable weights.

\textbf{Local Semantic Modulated Module.}
Slightly different from Eq.~(\ref{eq:12}), we generate specific modulated signals $\mathbf{a'}$ and $\bm{\eta'}$ for each capsule by calculating the contextual question representations of them. Afterwards, the similar computational procedure is formulated as follows,
\begin{equation}
    \label{eq:13}
    \left\{ \begin{array}{ll} 
    \mathcal{R}_{4}(\mathbf{V}_{m}^{(t)}) = \mathbf{a}' \odot \frac{\mathbf{V}^{(t)}_{m}-\mu(\mathbf{V}^{(t)}_{m})}{\sigma(\mathbf{V}^{(t)}_{m})} + \bm{\eta}', \\
    \mathbf{a}' = \delta_{b}(\mathbf{W}_{a'}\mathbf{Q}^{(t)}_{v}+\mathbf{B}_{a'}), \\
    \bm{\eta}' = \delta_{b}(\mathbf{W}_{\eta'}\mathbf{Q}^{(t)}_{v}+\mathbf{B}_{\eta'}), \\
    \mathbf{Q}^{(t)}_{v}=\delta_{a}((\mathbf{W}_{4}\mathbf{Q})^\top (\mathbf{W}_{5}\mathbf{V}_{m}^{(t)}))\mathbf{Q},
\end{array} \right.
\end{equation}
where $\mathbf{W}_{x} \in \mathbb{R}^{d \times d} \ (x=4,5)$, $\mathbf{W}_{a'} \in \mathbb{R}^{d \times d}$, and $\mathbf{W}_{\eta'} \in \mathbb{R}^{d \times d}$ are learnable weights.

\subsubsection{Specialized Module for Knowledge-required VQA}
\

\textbf{Adaptive Knowledge-augmented Module.} This module is designed specifically for the knowledge-required VQA task. It targets at enhancing the knowledge representations of each capsule. Considering that the visual feature and the knowledge feature contribute distinctively to the answer for different questions, thus we formulate an adaptive knowledge-augmented strategy for capsule modulation. Specifically, we first measure the pairwise affinity between these capsules and knowledge features, and then depict the difference between the capsules and knowledge context of them as a gating control for modulation.
We summarize the above process as follows,
\begin{equation}
    \label{eq:14}
    \left\{ \begin{array}{ll} 
     \mathcal{R}_{5}(\mathbf{V}_{m}^{(t)}) = \mathbf{g}_{k} \odot \mathbf{K}_{v}^{(t)} + (1-\mathbf{g}_{k}) \odot \mathbf{V}_{m}^{(t)}, \\
    \mathbf{K}_{v}^{(t)} = \delta_{a}((\mathbf{W}_{6}\mathbf{K})^\top (\mathbf{W}_{7}\mathbf{V}_{m}^{(t)})) \mathbf{K}, \\ 
    \mathbf{g}_{k} = \delta_{b}(\mathbf{W}_{8}\mathbf{V}_{m}^{(t)}-\mathbf{W}_{9}  \mathbf{K}_{v}^{(t)}),
\end{array} \right.
\end{equation}
where $\mathbf{W}_{x} \in \mathbb{R}^{d \times d} \ (x=6,7,8,9)$ are learnable weights, and $\mathbf{g}_{k} \in \mathbb{R}^{d \times N}$ denotes the update gate determined by the mathematical difference between the knowledge and capsules.

\begin{table*}
    \centering
    \caption{Performance comparison on the $\textit{val}$, $\textit{test-dev}$, and $\textit{test-std}$ splits of VQA v2. The best two are highlighted in bold.}  
    \label{table:vqav2}  
    \scalebox{1.0}{
    \begin{tabular}{ccccccccccccc}
    	\toprule [1 pt]
    	\multirow{2}{*}{Methods} & \multicolumn{4}{c}{VQA v2 \ $\textit{val}$ \ (Acc. \%)} & \multicolumn{4}{c}{VQA v2 \ $\textit{test-dev}$ \ (Acc. \%)} & \multicolumn{4}{c}{VQA v2 \ $\textit{test-std}$ \ (Acc. \%)} \\
    	\cmidrule[1 pt](r){2-5} \cmidrule[1 pt](r){6-9} \cmidrule[1 pt](r){10-13} 
    	&  \textit{All}      &  \textit{Y/N}  &  \textit{Num.}  &  \textit{Other}
    	&  \textit{All}      &  \textit{Y/N}  &  \textit{Num.}  &  \textit{Other}
     	&  \textit{All}      &  \textit{Y/N}  &  \textit{Num.}  &  \textit{Other}  \\
    	\midrule[1 pt]
    	MCB~\cite{DBLP:conf/emnlp/FukuiPYRDR16}  & 60.45  & -  & - & -  & 62.27 & 78.82 & 38.28 & 53.36 & - & - & - & - \\
    	MLB~\cite{DBLP:conf/iclr/KimOLKHZ17} & 62.91 & - & - & - & 66.27 & 83.58 & 44.92 & 56.34 & 66.62 & - & - & -  \\
    	MUTAN~\cite{DBLP:conf/iccv/Ben-younesCCT17} & 63.61 & - & - & - & 66.01 & 82.88 & 44.54 & 56.50  & 66.38 & - & - & -  \\
    	BLOCK~\cite{DBLP:conf/aaai/Ben-younesCTC19} & - & - & - & - & 67.58 & 83.60 & 47.33 & 58.51 & 67.92 & 83.98 & 46.77 & 58.79  \\
    	\hline
    	CapsAtt~\cite{DBLP:conf/aaai/ZhouJSSC19} & - & - & - & - & - & - & - & - & 65.50 & 82.60 & 45.10 & 55.50   \\
    	UpDn~\cite{DBLP:conf/cvpr/00010BT0GZ18} & 63.15 & 80.07 & 42.87 & 55.81  & 65.32 & 81.82 & 44.21 & 56.05 & 65.67 & 82.20 & 43.90 & 56.26       \\ 
    	Counter~\cite{DBLP:conf/iclr/ZhangHP18} & - & - & - & -  & 68.09 & 83.14 & \textbf{51.62} & 58.97 & 68.41 & - & - & -            \\ 
    	DCN~\cite{DBLP:conf/cvpr/NguyenO18}  & 63.83 & - & - & - & 66.83 & 84.48 & 41.66 & 57.44 & 66.66 & 84.61 & 41.27 & 56.83    \\
    	$\mathrm{BAN}^{\dagger}$~\cite{DBLP:conf/nips/KimJZ18} & 65.18 & - & - & - & 68.16 & - & - & - & 68.59 & - & - & -  \\
        $\mathrm{MCAN}^{\dagger}$~\cite{DBLP:conf/cvpr/Yu0CT019} & \textbf{65.95} & \textbf{83.69} & 47.60 & \textbf{57.32} & \textbf{69.58} & \textbf{86.31} & \textbf{51.58} & \textbf{59.24} & \textbf{70.06} & \textbf{86.44} & \textbf{51.43} & \textbf{60.06} \\
        \hline
        N2NMN~\cite{DBLP:conf/iccv/HuARDS17} & - & - & - & -  & 64.70 & - & - & -  & - & - & - & -  \\
        StackNMN~\cite{DBLP:conf/eccv/HuADS18} & - & - & - & -  & 64.82 & - & - & -  & - & - & - & -  \\
    	$\mathrm{XNM}$-$\mathrm{Net}^{\dagger}$~\cite{DBLP:conf/cvpr/ShiZL19} & 64.78 & - & - & -  & 67.15 & - & - & -  & 67.56 & - & - & -  \\
    	$\mathrm{CoR}$-$\mathrm{3}^{\dagger}$~\cite{DBLP:conf/nips/WuLWD18} & 64.92 & - & - & - & 68.19 & 84.96 & 47.11 & 58.64 & 68.59 & 85.16 & 47.19 & 59.07       \\
    	SceneGCN~\cite{yang2019scene} & - & - & - & - & 66.81 & 82.72 & 46.85 & 57.77 & 67.14 & 83.16 & 46.61 & 57.89  \\
    	TRN+UpDn~\cite{DBLP:conf/eccv/HanWSZHT20} & 65.10 & 82.61 & 45.10 & 57.10 & 67.00 & 83.83  & 45.61 & 57.44 & 67.21 & - & - & -  \\
    	\hline
    	CGN~\cite{DBLP:conf/nips/Norcliffe-Brown18} & - & - & - & - & 66.45 & - & - & - & 66.18 & 82.91 & 47.13 & 56.22          \\
    	$\mathrm{ODA}^{\dagger}$~\cite{DBLP:conf/mm/WuLWD18}  & - & - & - & - & 68.17 & 84.66 & 48.04 & 58.68 & - & - & - & - \\
    	MuRel~\cite{DBLP:conf/cvpr/CadeneBCT19}   & 65.14 & 83.24 & \textbf{47.72} & 56.34 & 68.17 & 84.66 & 48.04 & 58.68 & 68.41 & - & - & - \\
    	ReGAT~\cite{DBLP:conf/iccv/LiGCL19} & 65.30 & - & - & - & - & - & - & - & - & - & - & -  \\
    	VCTREE+HL~\cite{DBLP:conf/cvpr/TangZWLL19} & 65.10 & 82.61 & 45.10 & 57.10 & 68.19 & 84.28 & 47.78 & 59.11 & 68.49 & 84.55 & 47.36 & 59.34  \\
		\textbf{Ours (w/o know.)}  & \textbf{66.59} & \textbf{85.15} & \textbf{48.27} & \textbf{58.07} & \textbf{69.23} & \textbf{85.42} & 51.16 & \textbf{59.35} & \textbf{69.69} & \textbf{86.11} & \textbf{50.88} & \textbf{60.15}   \\
    	\bottomrule[1 pt]
    \end{tabular}}
\end{table*}

\subsection{Answer Reasoning}
After several iterations of the SUPER layer, we derive the refined output capsules $\mathbf{U}^{(T)}$. We ultimately aggregate these capsules by the mean pooling operation whereby obtain the discriminative representation for the whole image. To further emphasize the semantic guidance of the question, the visual representation $\mathbf{U}^{(T)}$ is merged with the question embedding for final answer prediction. Formally, we summarize the above process as follows,
\begin{equation}
    \label{eq:15}
    \mathbf{y} = \delta_{b}(\mathbf{W}_{y} (\mathbf{W}_{u} \mathcal{P}(\mathbf{U}^{(T)}) + \mathbf{W}_{q}\mathcal{P}(\mathbf{Q}))),
\end{equation}
where $\mathcal{P}(\cdot)$ denotes the average pooling function, $\mathbf{W}_{y} \in \mathbb{R}^{A \times d}$, $\textbf{W}_{u} \in \mathbb{R}^{d \times d}$, and $\textbf{W}_{q} \in \mathbb{R}^{d \times d}$ are learnable weights, $\mathbf{y} \in \mathbb{R}^{A}$ is the answer distribution, and $A$ is the number of candidate answers. Following ~\cite{DBLP:conf/cvpr/00010BT0GZ18}, we use the binary cross-entropy loss function to train an answer classifier.

\section{Experiments}
\subsection{Experimental Settings} 
\subsubsection{Datasets}
We conducted extensive experiments on two categories of VQA datasets: conventional VQA datasets (i.e., VQA-v2, VQA-CP v2, and Visual7W) and knowledge-required VQA datasets (i.e., OK-VQA and FVQA). 

The two categories of VQA datasets are elaborated as follows. The conventional VQA datasets contain: 1) VQA v2~\cite{DBLP:conf/cvpr/GoyalKSBP17}, the most commonly used VQA benchmark dataset. The images are from the Microsoft COCO dataset~\cite{DBLP:conf/eccv/LinMBHPRDZ14}, and each question has ten answers from different annotators. According to the question types, the answers are accordingly divided into three categories: \textit{Yes/No}, \textit{Number}, and \textit{Other}. Beyond that, the dataset is split into \textit{train}, \textit{val}, and \textit{test} (or \textit{test-std}) splits, and 25 \% of the \textit{test-std} set is reserved as the \textit{test-dev} set. 2) VQA-CP v2, curated from the VQA v2 dataset, which is developed to evaluate the question-oriented bias reduction ability and the generalization ability in VQA models. Due to the significant difference of distribution between the train set and the test set, the VQA-CP v2 dataset is much more challenging than the VQA v2. And 3) Visual7W \textit{telling} ~\cite{DBLP:conf/cvpr/ZhuGBF16}, with 328K multi-choice visual questions of diverse types (\textit{What}, \textit{Where}, \textit{When}, \textit{Who}, \textit{Why}, and \textit{How}) based on 47K images, it is a dataset in the multi-choice setting with four candidate answers for each question. In particular, we formulated it as a multi-label classification problem and made the final choice by selecting the one with the highest probability.

The knowledge-required VQA datasets contain: 1) OK-VQA ~\cite{DBLP:conf/cvpr/MarinoRFM19}, the largest knowledge-based VQA dataset with more than 14K open-ended visual questions, which requires the open commonsense knowledge to reason and answer. The questions are divided into eleven categories: Vehicles and Transportation (VT); Brands, Companies, and Products (BCP); Objects, Material, and Clothing (OMC); Sports and Recreation (SR); Cooking and Food (CF); Geography, History, Language, and Culture (GHLC); People and Everyday Life (PEL); Plants and Animals (PA); Science and Technology (ST); Weather and Climate (WC); and Other. And 2) FVQA ~\cite{DBLP:journals/pami/WangWSDH18}, consisting of five data splits, each of which has more than 2K images and 5K questions, and each question-answer sample is accompanied by a piece of ground-truth fact retrieved from a given knowledge base. Note that we conduct experiments without using the given knowledge base or the provided ground-truth facts in the original FVQA. Instead, we attempt to leverage open knowledge to improve the knowledge-required VQA performance. Following ~\cite{DBLP:conf/mm/Li0020}, we did not compare with these fact retrieval based approaches~\cite{DBLP:conf/nips/NarasimhanLS18, DBLP:conf/eccv/NarasimhanS18}.

\subsubsection{Baselines}
To verify the effectiveness and generalization of our proposed SUPER in two VQA tasks, we compared it against the following models.
\begin{itemize}
    \item Methods that focus on multimodal embedding techniques, namely,
    LSTM+Att.~\cite{DBLP:conf/cvpr/ZhuGBF16},
    MCB~\cite{DBLP:conf/emnlp/FukuiPYRDR16},
    MLB~\cite{DBLP:conf/iclr/KimOLKHZ17},
    MUTAN~\cite{DBLP:conf/cvpr/WuNKRDGF18}, BLOCK~\cite{DBLP:conf/iccv/Ben-younesCCT17}, 
    AN~\cite{DBLP:conf/cvpr/MarinoRFM19}, 
    MUTAN+AN~\cite{DBLP:conf/cvpr/MarinoRFM19}, 
    KG-Aug~\cite{DBLP:conf/mm/Li0020}, 
    and MLB+KG-Aug~\cite{DBLP:conf/mm/Li0020};
    \item Methods that aim to design various deep attention-based network, namely, Up-Down~\cite{DBLP:conf/cvpr/00010BT0GZ18},
    DCN~\cite{DBLP:conf/cvpr/NguyenO18},
    BAN~\cite{DBLP:conf/nips/KimJZ18}, 
    MCAN~\cite{DBLP:conf/cvpr/Yu0CT019}, 
    BAN+AN~\cite{DBLP:conf/cvpr/MarinoRFM19}, and  BAN+KG-Aug~\cite{DBLP:conf/mm/Li0020};
    \item Methods that focus on visual relational modeling, namely,
    CGN~\cite{DBLP:conf/nips/Norcliffe-Brown18},
    MuRel~\cite{DBLP:conf/cvpr/CadeneBCT19},
    ReGAT~\cite{DBLP:conf/iccv/LiGCL19}, 
    ODA~\cite{DBLP:conf/mm/Peng0WWH19}, and
    Mucko~\cite{DBLP:conf/ijcai/ZhuYWS0W20};
    \item Methods that perform visual reasoning, N2NMN~\cite{DBLP:conf/iccv/HuARDS17}, 
    XNM-Net~\cite{DBLP:conf/cvpr/ShiZL19}, CoR-3~\cite{DBLP:conf/nips/WuLWD18}, TRN+UpDn~\cite{DBLP:conf/nips/WuLWD18}, and SceneGCN~\cite{yang2019scene}.
\end{itemize}
Note that we directly quoted the results of these baselines from their original papers except the ones marked by ``$\dagger$'', whose results are obtained by running their released code. In addition, for fair comparison, we disabled our \textit{adaptive knowledge-augmented module} when comparing with the methods that do not use knowledge. The notations ``w know.'' and ``w/o know.'' represent the model with knowledge and the model without knowledge, respectively.
\subsubsection{Evaluation Metric}
We adopted the standard VQA accuracy metric for evaluation~\cite{DBLP:conf/iccv/AntolALMBZP15}. Given an image and a corresponding question, for a predicted answer $a$, the accuracy is computed as follows,
\begin{equation}
    \label{eq:16}
    Acc_{a} = \mathrm{min}(1, \frac{\# \mathrm{humans \ that \ provide \ \textit{a}}}{3}).
\end{equation}

\begin{table*}[tp]
	\centering  
	\fontsize{7}{8}\selectfont
		\caption{Results comparison on the overall OK-VQA and each question category.} 
		\label{table:okvqa}  
		\scalebox{1.2}{
		\begin{tabular}{ccccccccccccc}
			\toprule[0.98 pt]
			\multirow{2}{*}{Method}&  
			\multicolumn{11}{c}{OK-VQA $\textit{test}$ \ (Acc. \%)}\cr  
			\cmidrule[0.5 pt](r){2-13}
			& \textit{Overall} & \textit{VT} & \textit{BCP} & \textit{OMC} 
			& \textit{SR} & \textit{CF} & \textit{GHLC} & \textit{PEL} & \textit{PA} & \textit{ST} & \textit{WC} & \textit{Other} \cr  
			\midrule[0.5 pt] 
			MLB~\cite{DBLP:conf/iclr/KimOLKHZ17}   
			& 20.40 & 20.00 & 16.24 & 17.86 & 24.61 & 22.39 & 17.67 & 17.62 & 20.67 & 18.49 & 31.27 & 17.55 \\
			BAN~\cite{DBLP:conf/nips/KimJZ18}  
			& 25.17 & 23.79 & 17.67 & 22.43 & 30.58 & 27.90 & 25.96 & 20.33 & 25.60 & 20.95 & 40.16 & 22.46 \\
			XNM-Net~\cite{DBLP:conf/cvpr/ShiZL19}            
			& 25.61 & 26.84 & 21.86 & 18.22 & 33.02 & 23.93 & 23.83 & 20.79 & 24.81 & 21.43 & 42.64 & 24.39 \\
			$\mathrm{MCAN}^{\dagger}$~\cite{DBLP:conf/cvpr/Yu0CT019}
			& 25.87 & 25.14 & 20.95 & 25.26 & 33.48 & 28.47 & 18.24 & 19.96 & 31.74 & 21.10 & 40.56 & 21.19  \\
		    \textbf{Ours (w/o know.)}     
			& \textbf{28.79} & \textbf{28.68} & \textbf{25.25} & \textbf{28.88} & \textbf{34.91} & \textbf{29.96} & \textbf{30.61} & \textbf{23.47} & \textbf{32.58} & \textbf{24.59} & \textbf{44.85} & \textbf{26.32}  \\
            \Xhline{1.2pt}
			BAN+AN~\cite{DBLP:conf/cvpr/MarinoRFM19}                 
			& 25.61 & 24.45 & 19.88 & 21.59 & 30.79 & 29.12 & 20.57 & 21.54 & 26.42 & 27.14 & 38.29 & 22.16  \\
			MUTAN+AN~\cite{DBLP:conf/cvpr/MarinoRFM19}               
			& 27.58 & 25.56 & 23.95 & 26.87 & 33.44 & 29.94 & 20.71 & 25.05 & 29.70 & 11.48 & 27.92 & 14.98 \\
			MLB+KG-Aug~\cite{DBLP:conf/mm/Li0020}          
			& 20.89 & 20.08 & 16.45 & 17.71 & 26.68 & 22.27 & 16.78 & 18.03 & 21.70 & 18.60 & 32.58 & 17.91  \\
			BAN+KG-Aug~\cite{DBLP:conf/mm/Li0020}             
			& 26.71 & 24.65 & 21.59 & 22.42 & 34.75 & 28.67 & 23.97 & 21.97 & 27.75 & 23.28 & 38.85 & 24.29 \\
			Mucko~\cite{DBLP:conf/ijcai/ZhuYWS0W20} & 29.20 & - & - & - & - & - & - & - & - & - & - & -  \\
			\textbf{Ours (w know.)}          
			& \textbf{30.46} & \textbf{26.92} & \textbf{26.13} & \textbf{29.46} & \textbf{35.83} & \textbf{30.86} & \textbf{31.93} & \textbf{24.78} & \textbf{33.69} & \textbf{27.94} & \textbf{45.70} & \textbf{27.47}  \\
			\bottomrule[1 pt] 
		\end{tabular} 
		}
    \vspace{-1em}
\end{table*}  
\subsection{Implementation Details}
For each image, we set a threshold for Faster RCNN to obtain $N$=36 object features. All questions are padded or truncated to the same length $L$=15. Each image-question pair has $K$=10 relevant knowledge facts. The dimensionality of $d_{v}$ and $d$ are set to 2048 and 512, respectively. The number of iterations of the SUPER layer $T \in \left\lbrace 1, 2, 4, 6, 8, 10 \right\rbrace $, and we experimentally set $T$ to 8 in our best report. To train the SUPER, we used the Adam optimizer~\cite{DBLP:journals/corr/KingmaB14} with $\beta_{1}=0.9$ and $\beta_{2}=0.98$. The base learning rate is set to $ \mathrm{min}(2.5te^{-5}, 1e^{-4}) $, where $t$ is the current epoch number starting from 1. After 16 epochs, the learning rate is decayed by 1/4 in every two epochs to $2.5te^{-5}$. The batch size is set to 64.
\vspace{-1em}
\subsection{Quantitative Analysis}
In this section, we compared our SUPER with the current state-of-the-art methods on five benchmark datasets. 
\subsubsection{Results on VQA v2}
Table~\ref{table:vqav2} shows the results trained on the \textit{train} (the left column) and \textit{train+val} (the middle and right column) splits of the VQA v2 dataset. We can observe that our SUPER model achieves the second best performance, substantially surpassing the most baselines. In particular, as the canonical neural module network, N2NMN and XNM-Net reason the visual scene by assembling a collections of function-specific modules at the mercy of the question-based layout generator. As can be seen from the table, our SUPER prominently outperforms them by 4.03 \% and 1.78 \% on the \textit{test-std} set, respectively, which demonstrates our superiority of the modular routing mechanism and powerful specialized modules over NMN-based models in terms of reasoning the real scene. Inspiringly, our method is comparable to the multi-layer attention-based model MCAN, wherein several self-attention and cross-attention layers are cascaded for deeper vision-semantic modeling. Beyond that, we are comparable to several formidable pretrained models in the VQA task, and the related comparison results can be found in Table~\ref{table:vqav2_with_pretrained_models}.
\subsubsection{Results on OK-VQA}
As shown in Table~\ref{table:okvqa}, we provided the comparision results of several conventional VQA baselines in the top rows, and knowledge-based baselines in the bottom rows. Compared with these two kinds of VQA baselines, our model achieves the best performance among all models. Specifically, Mucko~\cite{DBLP:conf/ijcai/ZhuYWS0W20} adopts a modality-aware heterogeneous stacked graph convolutional network for complementary evidence exploration. This model employs the static pre-defined graph network for relational reasoning, which may sufficiently exploit more reasoning patterns. Nevertheless, with fewer parameters, our model outperforms Mucko by 0.86\%, which demonstrates the potential of our semantic-aware modular routing rationale towards flexible and delicate reasoning.
\subsubsection{Results on FVQA}
The results are reported in Table~\ref{table:fvqa}. It is worth noting that our model provides a substantial gain over the best knowledge-based baseline BAN + KG-Aug by 9.17\%. These results verify the effectiveness of our iterative SUPER layer, especially the well-designed \textit{adaptive knowledge-augmented module}.

\subsubsection{Results on VQA-CP v2}
To demonstrate the generalizability of our SUPER model, we also conducted experiments on the VQA-CP v2 dataset~\cite{DBLP:conf/nips/RamakrishnanAL18}. The corresponding results are reported in Table~\ref{table:performance_comparison_vqacp}. GVQA~\cite{DBLP:conf/cvpr/AgrawalBPK18}, CSS~\cite{DBLP:conf/emnlp/LiangJHZ20} and AdvReg.~\cite{DBLP:conf/nips/RamakrishnanAL18} are all the de-biased models, which are designed for alleviating the language prior problem. Although our proposed method is not specialized to tackle this problem, it provides a substantial gain over the GVQA model by 10.48\%. In addition, DC-GCN~\cite{DBLP:conf/acl/HuangWCZCLL20} supports intra- and inter-modality deep interactions, where its dependency parsing technique can effectively diminish the problem of question-based overfitting. We observe that our SUPER model still surpasses it by a slight margin. Strikingly, SUPER achieves a clear-cut improvement over all the competitors on the ``\textit{other}'' category, which contains more questions that involve semantic awareness and external knowledge, such as ``\textit{why}'' and ``\textit{how}''. This observation conforms to that our SUPER possesses powerful reasoning capacity and strong generalizabitlity.
\subsubsection{Results on Visual7W telling}
Table~\ref{table:visual7w} summarizes the accuracy comparison results. Note that, as to the Visual7W \textit{telling} dataset with large bias~\cite{DBLP:conf/eccv/JabriJM16}, MLP~\cite{DBLP:conf/cvpr/ZhuGBF16}, KDMN~\cite{DBLP:journals/corr/abs-1712-00733}, Up-Down~\cite{DBLP:conf/cvpr/00010BT0GZ18}, and BAN~\cite{DBLP:conf/nips/KimJZ18} all compete surprisingly ``well'' with the strong baseline MCAN~\cite{DBLP:conf/cvpr/Yu0CT019} and our SUPER. As concluded in~\cite{DBLP:conf/eccv/JabriJM16}, a model that appears qualitatively better on this dataset (i.e., Visual7W $\textit{telling}$) may perform worse quantitatively, because it captures dataset biases less well. On the contrary, the best-performing models are those who can exploit biases in this dataset the best, i.e., models that “cheat” the best. These observations demonstrate that our SUPER can be immune to the underlying biases of dataset by learning the flexible routing structure.



\begin{figure*}
 \hspace{-3mm}
 \includegraphics[width=180mm]{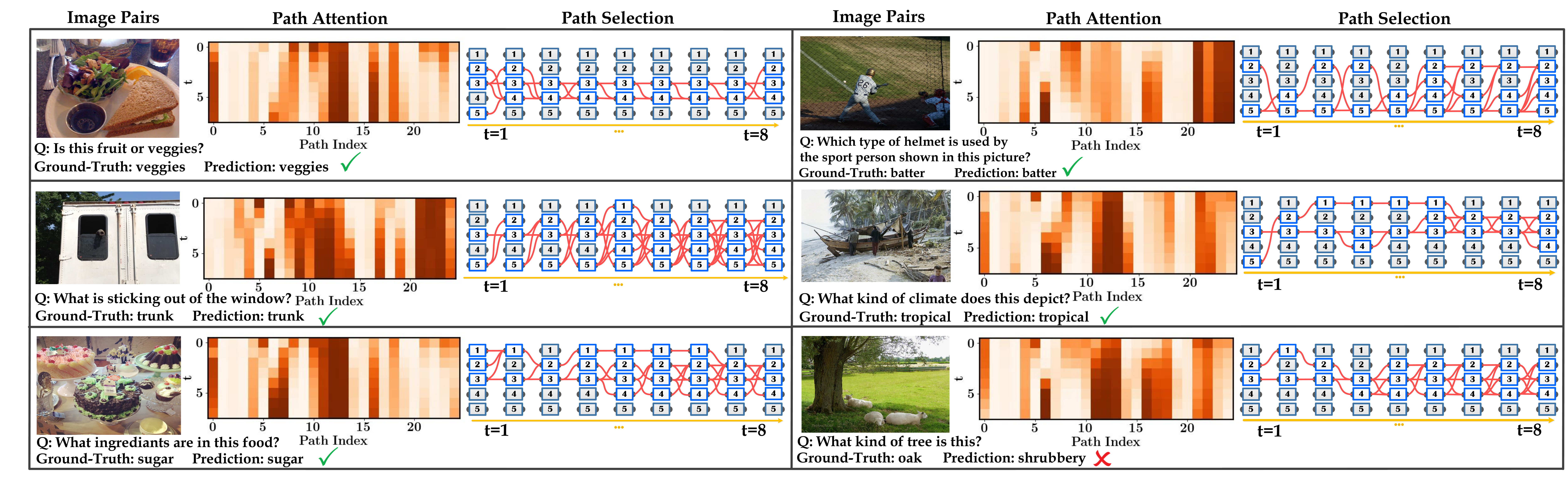}
 \vspace{-0.5em}
 \caption{Visualization of the capsule modulation patterns. We only show the semantic-aware modular routing in the SUPER layer for brevity, and the path index within [0-25] represents the number of routes densely constructed by these modules. The red paths denote the activated routes in each iteration.}
 \vspace{-1.1em}
 \label{fig:path_visualization}
\end{figure*}
\subsection{Ablation Study}
To get deeper insights into our model, we conducted some ablation experiments on the OK-VQA dataset to verify the effectiveness of each component in our SUPER layer.
\subsubsection{Semantic-aware Modular Routing}
In this section, we measured the effect of each module and router by evaluating the model accuracy when specific part gets removed or replaced. As reported in Table~\ref{table:ablation_analysis}, we observed a significant accuracy decrease when removing $\mathcal{R}_{1}$, which verifies that $\mathcal{R}_{1}$ is indispensable in capturing the delicate context information between capsules for relational reasoning. Besides, our model achieves better results than w/o $\mathcal{R}_{3}$ and w/o $\mathcal{R}_{4}$, which reveals that both global and local question information is pivotal to strengthen the semantic representations of capsule. Moreover, the large performance drop of w/o $\mathcal{R}_{5}$ indicates that our commonsense knowledge can largely facilitate the knowledge reasoning. In general, our proposed model largely exceeds all variants on both conventional VQA and knowledge-required VQA, verifying the effectiveness and complementarity of the five specialized modules. Pertaining to the router, we introduced two variants to validate the superiority of our soft router: 1) $\textbf{r.w. \ Random Router}$, replace our router with a random router. We independently derived the path probability of each module from a uniform distribution (i.e., $g_{m \rightarrow m^{*}} \in U(0, 1)$); and
2) $\textbf{w/o \ Router}$, removing the router. In other words, the specialized module directly delivers the output to other modules without control gate. As can be seen from Table~\ref{table:ablation_analysis}, our proposed router achieves the best performance over these two variants consistently. This is primarily due to the fact that module selection arguably benefits some merits from the data-driven mechanism. If the modules are entirely or randomly activated \footnote{More information flows from other modules to this module.}, some redundant information may be introduced by the irrelevant modules, thereby leading to suboptimal results.
\subsubsection{Gating Agreements}
Removing \textit{gating agreements} leads to a performance drop by 0.81\%. It indicates that focusing on considerable capsules is more beneficial to model decisions.
\subsubsection{Memory Reactivation}
To assess the impact of \textit{memory reactivation}, we removed it to derive a variant. The performance degrades dramatically by 1.53\%. This reveals that this component could preserve and rectify the discriminative information whereby boost the VQA performance.

\begin{table}[tp]
	\centering  
	\fontsize{7.3}{8}\selectfont
		\caption{Performance evaluated on FVQA.}
		\label{table:fvqa}  
		\scalebox{0.89}{
		\begin{tabular}{ccccccc}  
			\toprule[0.98 pt]
			\multirow{2}{*}{Method}&  
			\multicolumn{6}{c}{FVQA $\textit{val}$ \ (Acc. \%)}\cr  
			\cmidrule[0.5 pt](lr){2-7}
			& \textit{Average} & \textit{Split-1} & \textit{Split-2} & \textit{Split-3} & \textit{Split-4} & \textit{Split-5} \cr  
			\midrule[0.5 pt] 
			$\mathrm{LSTM}$+$\mathrm{Att.}^{\dagger}$~\cite{DBLP:conf/cvpr/ZhuGBF16}  & 24.26  & 24.02  & 22.78  & 23.65 & 25.71  & 25.14 \\
 			$\mathrm{HieCoAtt}^{\dagger}$~\cite{DBLP:conf/nips/LuYBP16}  & 32.42  & 31.98  & 31.24  & 32.14 & 33.58  & 33.17 \\
            $\mathrm{BAN}^{\dagger}$~\cite{DBLP:conf/nips/KimJZ18} & 35.69  & -  & -  & - & -  & - \\
            $\mathrm{MCAN}^{\dagger}$~\cite{DBLP:conf/cvpr/Yu0CT019}  & 44.48  & 43.21  & 41.89  & 43.19 & 47.51  & 46.64 \\
            \textbf{Ours (w/o know.)} & \textbf{47.75} & \textbf{47.34} & \textbf{46.62} & \textbf{47.08} & \textbf{49.16} & \textbf{48.56} \\
            \hline
            KG-Aug~\cite{DBLP:conf/mm/Li0020} & 31.96  & -  & -  & - & -  & - \\
            BAN + KG-Aug~\cite{DBLP:conf/mm/Li0020} & 38.58  & - & - & -  & - & - \\
            \textbf{Ours (w know.)} & \textbf{48.90}  & \textbf{48.98}  &  \textbf{47.24} &  \textbf{48.96} & \textbf{50.07} & \textbf{49.28}\\
			\bottomrule[1 pt] 
		\end{tabular} 
		}
\end{table}  

\begin{table}[tp]  
 	\centering  
 	\fontsize{7}{8}\selectfont  
 		\caption{Performance comparison on the $\textit{test-dev}$ and $\textit{test-std}$ splits of the VQA v2 dataset. Our model is trained on the \textit{train} + \textit{val} splits with extra Visual Genome augmentation~\cite{DBLP:journals/ijcv/KrishnaZGJHKCKL17}. The symbol ``*'' refers to the models that are pretrained on the large corpus \textit{Conceptual Captions}~\cite{sharma-etal-2018-conceptual} or $\textit{COCO captions}$  ~\cite{DBLP:conf/eccv/LinMBHPRDZ14}, and ``$\dagger$'' indicates the models without pretraining.}  
  		\label{table:vqav2_with_pretrained_models} 
  		\scalebox{1}{
 		\begin{tabular}{p{20mm}<{\centering}p{10mm}<{\centering}p{10mm}<{\centering}p{17mm}<{\centering}}  
 			\toprule[1 pt]
 			\multirow{2}{*}{Methods}&  
 			\multicolumn{2}{c}{VQA v2 (Acc. \%)} & \multirow{2}{*}{Param. Size}\cr  
 			\cmidrule[0.5 pt](lr){2-3} 
 			& $\textit{test-dev}$ & $\textit{test-std}$  \cr  
 			\midrule[0.5 pt] 
 			$\mathrm{VL}$-$\mathrm{BERT}^{\dagger}$~\cite{DBLP:conf/iclr/SuZCLLWD20} & 69.58 & - & 1331 M      \\
 			$\mathrm{ViLBERT}^{\dagger}$~\cite{DBLP:conf/nips/LuBPL19} & 68.93 & - & 954 M   \\
 			$\mathrm{VisualBERT}^{\dagger}$~\cite{li2019visualbert} & 70.18 & - & 428 M       \\
 			\hline
 			$\mathrm{VL}$-$\mathrm{BERT}^{*}$~\cite{DBLP:conf/iclr/SuZCLLWD20} & \textbf{71.16} & - & 1331 M     \\ 
 			$\mathrm{ViLBERT}^{*}$~\cite{DBLP:conf/nips/LuBPL19} & 70.55 & 70.92 & 954 M  \\ 
 			$\mathrm{VisualBERT}^{*}$~\cite{li2019visualbert} & 70.80 & \textbf{71.00} & 428 M    \\
 			\hline
 			$\textbf{Ours (w/o know.)}$   & 69.56  & 70.36 & 109.31 M \\
 			\bottomrule[1 pt] 
 		\end{tabular}
 		}
\end{table}

\begin{table}[tp]
	\centering  
	\fontsize{7}{8}\selectfont
		\caption{Performance evaluated on the $\textit{test}$ split of the VQA-CP v2 dataset. Models with ``*'' are specially designed for solving the language bias problem~\cite{DBLP:conf/cvpr/AgrawalBPK18}.}
		\label{table:performance_comparison_vqacp}  
		\scalebox{0.95}{
		\begin{tabular}{p{24mm}<{\centering}p{10mm}<{\centering}p{10mm}<{\centering}p{10mm}<{\centering}p{10mm}<{\centering}}  
			\toprule[0.98 pt]
			\multirow{2}{*}{Methods}&  
			\multicolumn{4}{c}{VQA-CP v2 $\textit{test}$ \ (Acc. \%)}\cr  
			\cmidrule[0.5 pt](lr){2-5}
			& \textit{All} & \textit{Y/N} & \textit{Num.} & \textit{Other} \cr  
			\midrule[0.5 pt] 
			 MCB~\cite{DBLP:conf/iccv/YuY0T17}  & 36.33 & 41.01 & 11.96 & 25.72 \\
			 UpDn~\cite{DBLP:conf/cvpr/00010BT0GZ18} & 39.74 & 42.70 & 11.93 & 46.05\\
		     BAN~\cite{DBLP:conf/nips/KimJZ18} & 39.31 & - & - & - \\
		     MuRel~\cite{DBLP:conf/nips/WuLWD18} & 39.54 & 42.85 & 13.17 & 45.05\\
		     ReGAT~\cite{DBLP:conf/iccv/LiGCL19} & 40.42 & - & - & -\\
		     MRA-Net~\cite{tpamipeng20} & 40.45 & 44.53 & 13.05 & 45.83\\
		     DC-GCN~\cite{DBLP:conf/acl/HuangWCZCLL20} & 41.47 & - & - & -\\
			\midrule[0.5 pt]  
	         $\mathrm{GVQA}^{*}$~\cite{DBLP:conf/cvpr/AgrawalBPK18} & 31.30 & 57.99 & 13.68 & 22.14 \\
	         $\mathrm{CSS}^{*}$~\cite{DBLP:conf/emnlp/LiangJHZ20} & 41.16 & 43.96 & 12.78 & 47.78 \\
	         $\mathrm{AdvReg.}^{*}$~\cite{DBLP:conf/nips/RamakrishnanAL18} & 41.17 & \textbf{65.49} & 15.48 & 35.48\\
		    \midrule [1 pt]
		    \textbf{Ours (w/o know.)} & \textbf{41.78} & 46.54 & \textbf{15.76} & \textbf{49.25} \\
			\bottomrule[1 pt] 
		\end{tabular} 
		}
\end{table}  

\begin{table}[tp]
	\centering  
	\fontsize{7.5}{8}\selectfont
		\caption{Performance evaluated on Visual7W.}
		\label{table:visual7w}  
		\scalebox{0.86}{
		\begin{tabular}{cccccccc}  
			\toprule[0.98 pt]
			\multirow{2}{*}{Method}&  
			\multicolumn{7}{c}{Visual7W $\textit{telling}$ \ (Acc. \%)}\cr  
			\cmidrule[0.5 pt](lr){2-8}
			& \textit{Overall} & \textit{What} & \textit{Where} & \textit{When}
			& \textit{Who} & \textit{Why} & \textit{How} \cr
			\midrule[0.5 pt] 
			
			LSTM+Att.~\cite{DBLP:conf/cvpr/ZhuGBF16}
			& 54.30 & 51.50 & 57.00 & 75.00 & 59.50 & 55.50 & 49.80 \\
            MLP~\cite{DBLP:conf/cvpr/ZhuGBF16}
            & 67.10 & 64.50 & 75.90 & 82.10 & 72.90 & 68.00 & 56.40  \\
			MCB~\cite{DBLP:conf/emnlp/FukuiPYRDR16}        
			& 62.20  & 60.30  & 70.40 & 79.50 & 69.20  & 58.20  & 51.10  \\
			KDMN~\cite{DBLP:journals/corr/abs-1712-00733}
			& 66.00 & 64.60 & 73.10 & 81.30 & 73.90 & 64.10 & 53.30 \\
            $\mathrm{Up}$-$\mathrm{Down}^{\dagger}$~\cite{DBLP:conf/cvpr/00010BT0GZ18}
			& 66.82 & 65.78 & 74.28 & 81.97 & 75.40 & 64.27 & 58.64 \\
			$\mathrm{BAN}^{\dagger}$~\cite{DBLP:conf/nips/KimJZ18}   
            & 70.79 & 70.23 & 76.52 & 82.06 & 78.45 & 64.76 & 57.68 \\
            $\mathrm{MCAN}^{\dagger}$~\cite{DBLP:conf/cvpr/Yu0CT019} 
			& \textbf{63.68} & \textbf{62.59} & \textbf{71.68} & \textbf{79.85} & \textbf{72.36} & \textbf{60.41} & \textbf{54.53}  \\
			\hline
	        \textbf{Ours (w/o know.)} & \textbf{64.07} & \textbf{62.64} & \textbf{72.34} & \textbf{79.87} & \textbf{72.96} & \textbf{59.98} & \textbf{53.78}  \\
			\bottomrule[1 pt] 
		\end{tabular} 
		} 
		\vspace{-1em}
\end{table}  


\begin{table}[tp]  
	\centering  
	\fontsize{7}{8}\selectfont  
		\caption{Ablation study results on OK-VQA.}
		\vspace{-1em}
		\label{table:ablation_analysis}  
		\scalebox{1.1}{
		\begin{tabular}{p{27mm}<{\centering}p{11mm}<{\centering}p{11mm}<{\centering}}  
			\toprule[1 pt]
			\multirow{2}{*}{Method}&  
			\multicolumn{2}{c}{OK-VQA $\textit{test}$ (Acc. \%)}\cr  
			\cmidrule[0.5 pt](lr){2-3}
			& \textit{All} & -$\Delta$ $\downarrow$ \cr  
			\midrule[0.5 pt] 
		    Full Model & \textbf{30.46}&\textbf{0.00}\cr
			\midrule[0.5 pt] 
			\multicolumn{3}{c}{(Five Specialized Modules)} \cr
			w/o $\mathcal{R}_{1}$ & 29.27    & -1.19  \cr
			w/o $\mathcal{R}_{2}$  & 28.85   & -1.61 \cr
			w/o $\mathcal{R}_{3}$ & 28.90   & -1.56  \cr
 			w/o $\mathcal{R}_{4}$  & 28.58   & -1.88  \cr
 			w/o $\mathcal{R}_{5}$ & 28.79    & -1.67  \cr 
 			\midrule[0.5 pt]  
 			\multicolumn{3}{c}{(Router)} \cr
			w/o Router & 27.79 & -2.67 \cr
 			r.w. Random Router & 27.12 & -3.34 \cr
		    \midrule[0.5 pt]  
            w/o Gating Agreements & 29.25    & -1.21  \cr
			\midrule[0.5 pt]  
            w/o Memory Reactivation & 28.53   & -1.93  \cr
			\bottomrule[1 pt] 
		\end{tabular} 
   }
   	\vspace{-2.5em}
\end{table}  
\subsection{Parameter Analysis}
\subsubsection{The Effect of the Number of Iterations $T$} We carried out experiments on the OK-VQA dataset to explore how the number of iterations $T$ of the SUPER layer affects the model performance. As shown in Table~\ref{table:okvqa-iterations}, we observed that increasing $T$ ($T \leq 8$) brings a consistent gain in the overall accuracy. It indicates that understanding complicated visual scene requires more sophisticated reasoning routes, which can be fulfilled via more iterations. However, when $T$ is greater than 8, the performance begins to drop. This phenomenon can be explained by that the exploration of routing space gradually tends to be saturated. In this case, increasing $T$ may lead to unstable gradients during training and make the optimization difficult.
\subsubsection{The Model Size} 
Notably, our SUPER could ingeniously balance the performance and parameter size, as shown in Table~\ref{table:model_size}. The first two (i.e., MLB and MCB) are embedding-based models, and the middle two (i.e., BAN and MCAN) are multi-layer attention-based stacked models. Specifically, comparing with MCAN, the parameter size of SUPER is only one-third, whereas SUPER demonstrates clear-cut improvement over MCAN on three datasets, namely, Visual7W, OK-VQA, and FVQA. 
\begin{table}
	\centering  
	\fontsize{7.1}{8}\selectfont
		\caption{Performance with different number of iterations $T$ on OK-VQA.}
		\label{table:okvqa-iterations}  
		\scalebox{1.05}{
		\begin{tabular}{cccccccc}  
			\toprule[0.98 pt]
			$T$ & 1 & 2 & 4 & 6 & 8 & 10 & 12 \\
			\hline
			Acc. (\%) & 28.29  & 28.52  & 28.98 & 29.20 & \textbf{30.46} & 28.86 & 28.46\\
			\bottomrule[1 pt] 
		\end{tabular} 
		}
\end{table}  

\begin{table}
	\centering  
	\fontsize{7}{8}\selectfont
		\caption{Comparisons of the parameter sizes (M) of different models.}
		\label{table:model_size}  
		\scalebox{0.93}{
		\begin{tabular}{ccccccc}  
			\toprule[0.98 pt]
			Methods & MLB & MCB & BAN & MCAN & \textbf{Ours (w/o know.)} \\
			\hline
			Param. Size & 53.90 M & 70.56 M & 120.22 M & 317.70 M & \textbf{109.31} M\\
			\bottomrule[1 pt] 
		\end{tabular} 
		}
\end{table}  

\subsection{Qualitative Results} 
To obtain deeper insights into our semantic-aware modular routing process, we visualized several results with different paths. Concretely, we employed 0.6 as the threshold to discretize the learned paths (i.e., we only show the paths that the probability values are greater than pre-defined threshold) for improving the intuitiveness. As shown in Figure~\ref{fig:path_visualization}, we have the following observations: 1) Simpler inputs tend to activate less paths. For example, the first question in the first row simply asks about the type of food, and it is inclined to activate only two types of specialized modules for visual modulation, which may not need much fine-grained information; and 2) knowledge-based visual questions require more \textit{adaptive knowledge-augmented module} and \textit{local semantic modulated module} since they necessitate more external knowledge for facilitating the fine-grained vision-semantic understanding. Taking the left second question as an example, when asked the ingredients of food, the model first attends the ``cake'' via \textit{local semantic modulated module} and leverages \textit{focal context-aware enhanced module} to gradually reason the whole visual scene. Afterwards, \textit{adaptive knowledge-augmented module} is activated to incorporate the facts (sugar, Isa, ingredient) and (cake, RelatedTo, sugar) into the reasoning process, for delicate knowledge reasoning. 

\begin{figure}
 \centering
 \includegraphics[width=80.5mm]{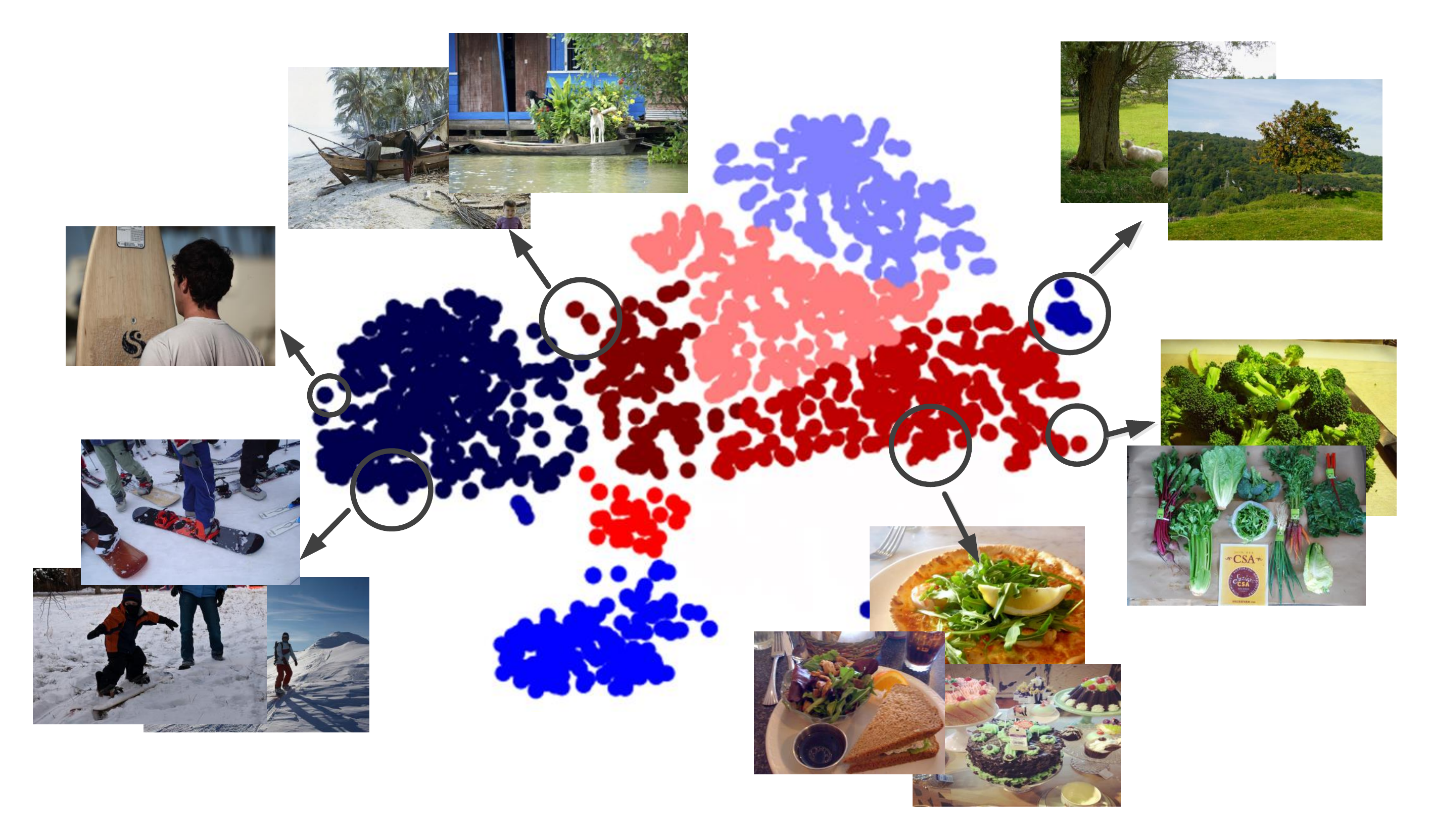}
  \vspace{-0.5em}
 \caption{Visualization of the learned path vectors using T-SNE on the OK-VQA dataset.}
 \label{fig:tsne}
 \vspace{-1em}
\end{figure}

Apart from achieving the promising performance with fewer parameters, the key 
property of SUPER is that it could adaptively activate different reasoning paths for diverse inputs by our shared SUPER layer. We thus exhibited some images and visualized their path vectors learned in \textit{semantic-aware modular routing}. Specifically, we concatenated all the path attention learned in each iteration into one path vector, and then used the t-SNE~\cite{tsne} algorithm to map the path vector into a two-dimensional Euclid space. Afterwards, we clustered these 2D vectors into 8 groups in 8 colors. As shown in Figure~\ref{fig:tsne}, we observed that the images related to the tree (i.e., the points marked in navy blue, in the rightmost of Figure~\ref{fig:tsne}) and the ones related to vegetables (i.e., the points marked in carmine) can be well distinguished. These two groups are both related to green plants, however they are still wide apart. Because there is much difference in their fine-grained context semantics. Moreover, there is a large margin between the carmine points and the dark blue ones (associated with skiing), as they are clearly distinct. Our proposed data-dependent router is able to transfer these diverse semantic information to the routes selection. These results reveal that our model can flexibly activate specific paths for distinct inputs, therefore the distribution of learned paths is arguably consistent with that of the vision-semantic information.

\section {Conclusion and Future Work} 
In this paper, we present a dynamic modular routing framework for VQA, wherein three well-devised components systematically work as a whole and reciprocate each other, for flexible routing exploration and powerful vision-semantic representation calibration. Extensive experiments on five benchmarks validate strong generalizability as well as satisfactory parametric efficiency of our proposed model.

In the future, we plan to explore more applications of the modular routing mechanism in various multimodal learning tasks or large-scale pretrained tasks, making it more flexible and extensible.


\ifCLASSOPTIONcaptionsoff
  \newpage
\fi

\small
\bibliographystyle{IEEEtran}
\bibliography{tip}

\vspace{-4.0em}
\begin{IEEEbiography}	[{\includegraphics[width=1in,height=1.25in,clip,keepaspectratio]{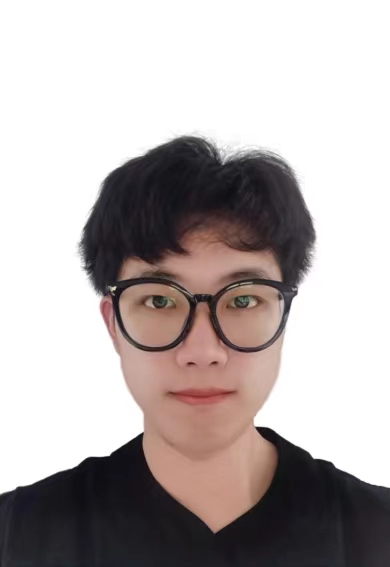}}]{Yudong Han}  received the B.E. degree from Shandong Normal University in 2020. He is currently pursuing the master’s degree with the School of Computing Science and Technology, Shandong University. His research interests include multimedia
computing and image processing. He has served as a Reviewer for various conferences and journals, such as ACM MM, AAAI, and IEEE TCYB.
\end{IEEEbiography}
\vspace{-4.25em}
\begin{IEEEbiography}
	[{\includegraphics[width=1in,height=1.25in,clip,keepaspectratio]{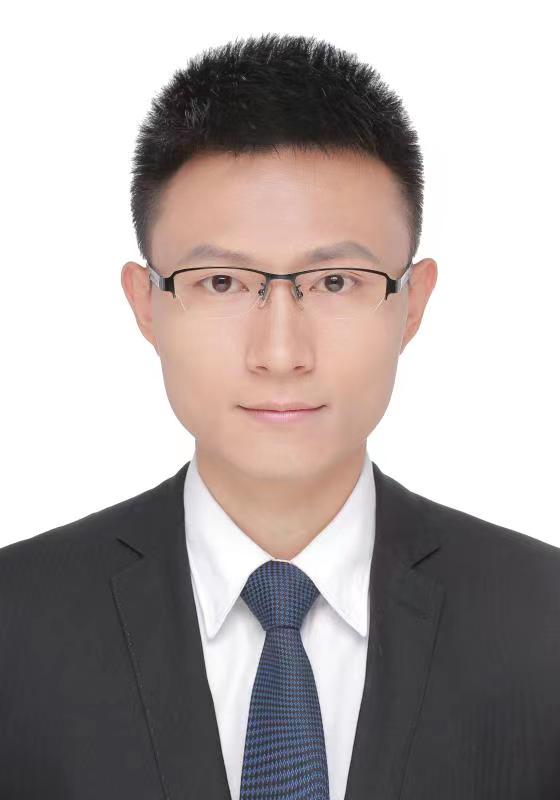}}]{Jianhua Yin} received the Ph.D. degree in computer science and technology from Tsinghua University, Beijing, China, in 2017. He is currently an Associate Professor with the School of Computer Science and Technology, Shandong University, Jinan, China. He has published several papers in the top venues, such as ACM TOIS, IEEE TMM, ACM MM, ACM SIGKDD, and ACM SIGIR. His research interests include data mining and machine learning applications.
\end{IEEEbiography}
\vspace{-4.2em}
\begin{IEEEbiography}
	[{\includegraphics[width=1in,height=1.25in,clip,keepaspectratio]{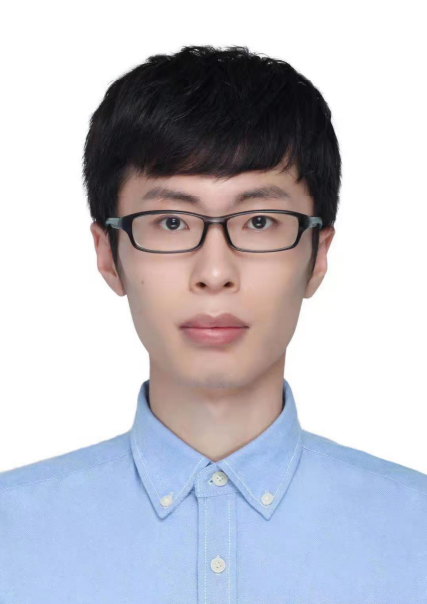}}]{Jianlong Wu} (Member, IEEE) received his B.Eng. and Ph.D. degrees from Huazhong University of Science and Technology in 2014 and Peking University in 2019, respectively. He is currently an assistant professor with the School of Computer Science and Technology, Shandong University. His research interests lie primarily in computer vision and machine learning. He has published more than 30 research papers in top journals and conferences, such as TIP, ICML, NeurIPS, and ICCV. He received many awards, such as outstanding reviewer of ICML 2020,  and the Best Student Paper of SIGIR 2021. He serves as a Senior Program Committee Member of IJCAI 2021, an area chair of ICPR 2022/2020, and a reviewer for many top journals and conferences, including TPAMI, IJCV, ICML, and ICCV.
\end{IEEEbiography}
\vspace{-4.3em}
\begin{IEEEbiography}
	[{\includegraphics[width=1in,height=1.25in,clip,keepaspectratio]{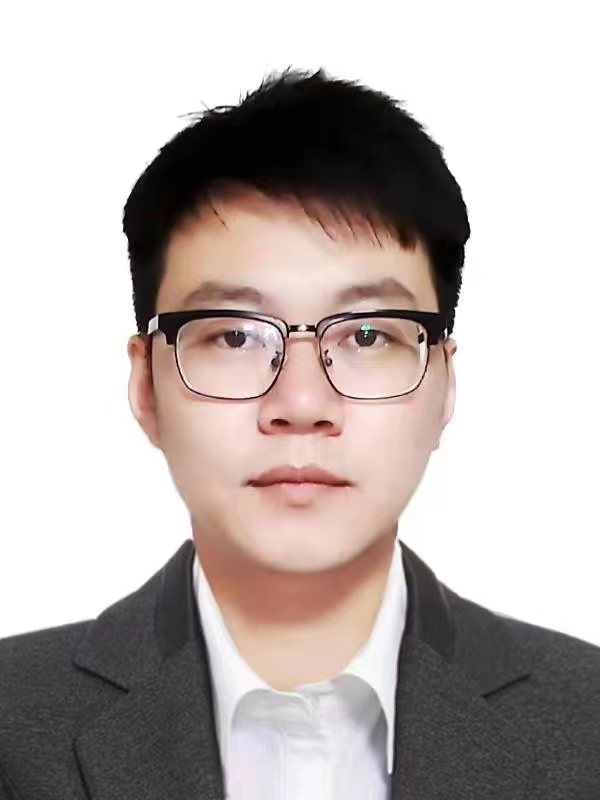}}]{Yinwei Wei} is currently a research fellow with NExT++, National University of Singapore. He received his MS degree from Tianjin University and Ph.D. degree from Shandong University, respectively. His research interests include multimedia computing and recommendation. Several works have been published in top forums, such as ACM MM, SIGIR, TIP, TKDE, and TMM. Moreover, Dr. Wei has served as the PC member for several conferences, such as MM, WSDM, AAAI, and IJCAI, and the reviewer for TPAMI, TKDE, and TIP.
\end{IEEEbiography}
\vspace{-4.3em}
\begin{IEEEbiography}
	[{\includegraphics[width=1in,height=1.25in,clip,keepaspectratio]{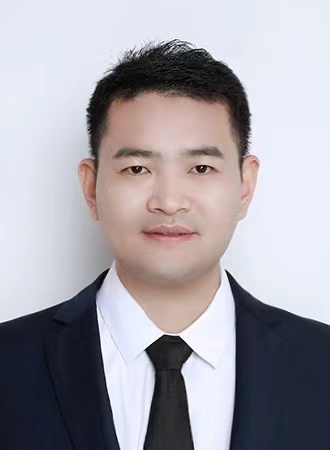}}]{Liqiang Nie} is currently the dean with the Department of Computer Science and Technology, Harbin Institute of Technology (Shenzhen). He received his B.Eng. and Ph.D. degree from Xi'an Jiaotong University and National University of Singapore (NUS), respectively. After PhD, Dr. Nie continued his research in NUS as a research fellow for three years. His research interests lie primarily in multimedia computing and information retrieval. Dr. Nie has co-/authored more than 100 papers and 4 books, received more than 14,000 Google Scholar citations. He is an AE of IEEE TKDE, IEEE TMM, IEEE TCSVT, ACM ToMM, and Information Science. Meanwhile, he is the regular area chair of ACM MM, NeurIPS, IJCAI and AAAI. He is a member of ICME steering committee. He has received many awards, like ACM MM and SIGIR best paper honorable mention in 2019, SIGMM rising star in 2020, TR35 China 2020, DAMO Academy Young Fellow in 2020, and SIGIR best student paper in 2021.
\end{IEEEbiography}

\end{document}